%% file: main.tex
\documentclass[conference]{IEEEtran}
\pdfoutput=1

 \usepackage[pdftex]{graphicx}

\usepackage{subcaption}
\usepackage{balance}

\usepackage{url}
\usepackage{multirow}
\usepackage{adjustbox}

\begin{document}
%
\title{Recognizing Textures with Mobile Cameras for Pedestrian Safety Applications}
%
%
%
%

\author{\IEEEauthorblockN{Shubham Jain}
\IEEEauthorblockA{Department of Computer Science\\
Old Dominion University\\
Email: jain@cs.odu.edu}
\and
\IEEEauthorblockN{Marco Gruteser}
\IEEEauthorblockA{WINLAB\\
Rutgers University\\
Email: gruteser@winlab.rutgers.edu}
}

\markboth{IEEE Transactions on Mobile Computing}%
{IEEE Transactions on Mobile Computing}
\IEEEtitleabstractindextext{%
\begin{abstract}
As smartphone rooted distractions become commonplace, the lack of compelling safety measures has led to a rise in the number of injuries to distracted walkers. Various solutions address this problem by sensing a pedestrian's walking environment. Existing camera-based approaches have been largely limited to obstacle detection and other forms of object detection. Instead, we present TerraFirma, an approach that performs material recognition on the pedestrian's walking surface. We explore, first, how well commercial off-the-shelf smartphone cameras can learn texture to distinguish among paving materials in uncontrolled outdoor urban settings. 
Second, we aim at identifying when a distracted user is about to enter the street, which can be used to support safety functions such as warning the user to be cautious. 
To this end, we gather a unique dataset of street/sidewalk imagery from a pedestrian's perspective, that spans major cities like New York, Paris, and London.
We demonstrate that modern phone cameras  can be  enabled to distinguish materials of walking surfaces in urban areas with more than 90\% accuracy, and accurately identify when pedestrians transition from sidewalk to street.
\end{abstract}

}

\maketitle

\IEEEdisplaynontitleabstractindextext

%
\IEEEpeerreviewmaketitle

\input{introduction}

\input{background}

\input{challenges}

\input{method}

\input{experiments}

\input{evaluation}


\input{discussion}

\input{conclusion}
\balance

\bibliographystyle{IEEEtran}
\bibliography{ped_texture2,proposal_ped}
%

%




\vfill


\end{document}

%% file: introduction.tex
\section{Introduction}
\label{sec:introduction}

Mobile cameras have evolved over time, and can now support a multitude of techniques that seemed difficult less than a decade ago. Camera sensing is not only instrumental in the self-driving cars initiative~\cite{ny_selfdriving, onecam}, but also drone guidance~\cite{drone2, drone}, augmented reality~\cite{camera_ar}, real-time surveillance~\cite{panoptes} through dashboard mounted and body worn cameras~\cite{bodyworn}, agriculture IoT~\cite{farmbeats} and urban sensing~\cite{urbansensing1}.
Cameras are one of the richest sources of data, and ubiquitously deployed.

In the realm of pedestrian safety, mobile cameras, such as in-car cameras, are also increasingly used. Researchers have also explored the use of smartphone cameras to target distracted pedestrians~\cite{tactilepaving, walksafe}. Texting while walking is widely considered a safety risk.
On average, a pedestrian was killed in the United States every 2 hours and injured every 8 minutes, in 2014~\cite{nhtsa2014, bg}. While not all these accidents are related to distracted walking, it is notable that US pedestrian fatalities have risen in the past decade, to account for $15\%$ of all traffic fatalities. Research has attributed this increase to the phenomenon of distracted walking~\cite{osu}.

Existing camera-based pedestrian safety approaches, such as TypenWalk~\cite{typenwalk} uses the current camera view as a background for user's application, to help them observe their surroundings while using the application.
However, the onus to watch for hazards still lies with the pedestrian. This approach can help but it is not clear how many distracted pedestrians would pay adequate attention to the subtle screen background of the path in front. Walksafe~\cite{walksafe} is another smartphone camera based approach for pedestrians who talk while crossing the street. It detects approaching vehicles in direct line of sight using the rear camera. Although ingenuous, it can only detect vehicles when the phone is held up to the ear and only vehicles from one side. Outdoor obstacle detection~\cite{obstacledetection} using smartphones seeks to detect obstacles in the camera's frame that are potentially in the pedestrian's path. Recently, Tang et al~\cite{tactilepaving}, demonstrated the use of the smartphone rear camera for alerting distracted pedestrians by detecting tactile paving on sidewalks. We have observed that such paving, although desirable, is not commonplace, which makes the approach hard to deploy in diverse environments. 
Despite its progress, camera sensing in the real world has been largely limited to object detection and recognition. Most prior research mentioned above applies object recognition techniques. 

The question that arises is: {\it Can we  enable mobile cameras to sense the environment without necessitating the presence of specific objects?} 
While several such techniques have been demonstrated on sophisticated camera setups~\cite{curet} and controlled environments~\cite{drexel}, it is unclear whether they work with the smaller, lower quality smartphone image sensors and lenses.
One such technique is that of recognizing and identifying material. Differentiating materials is harder than recognizing objects, even for the human eyes. 
Objects have well-defined shapes and high level attributes that can be quantified to identify them against a background, even in extreme lighting and weather conditions~\cite{obj_lighting, obj_weather}. On the other hand, texture attributes used for material recognition are a quantitative measure of the local and global arrangement of individual pixels. It is unclear how the smaller sensors on smartphone cameras capture this spatial relationship between pixels. Noise in camera pixels are generated due to the photons from ambient lighting. At the output of a camera, the noise current in each camera pixel manifests as fluctuations in the intensity of that pixel~\cite{vmimo}. Mobile camera use in outdoor environments also leads to much more significant image quality degradation than those in staged indoor environments---for example, due to  motion blur, over and under exposure, and compression artifacts.

\begin{figure}
\centering
\begin{subfigure}[t]{.35\linewidth}
\includegraphics[width=\linewidth]{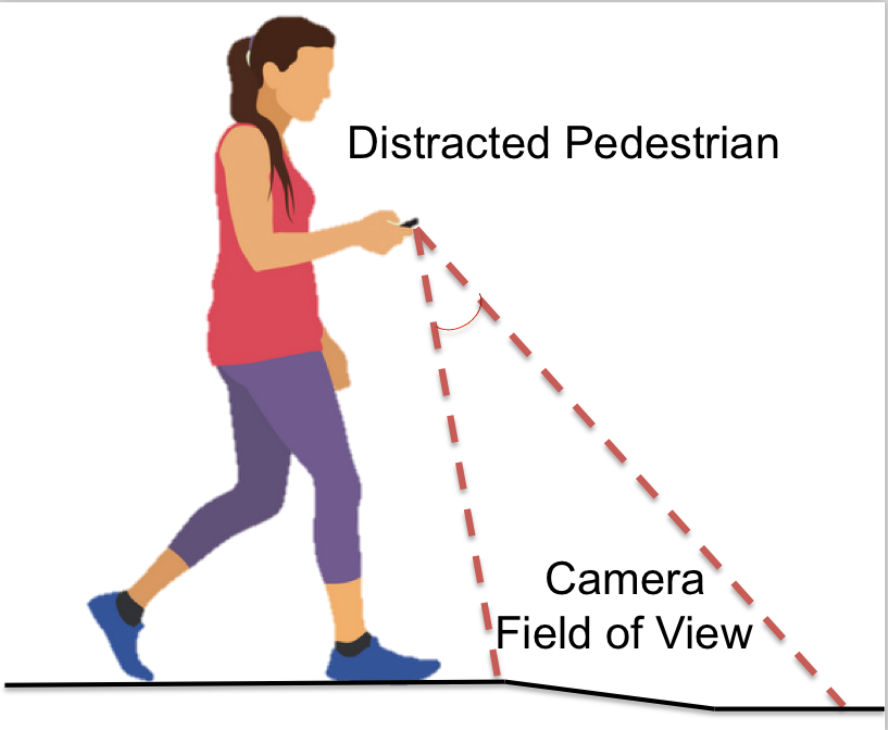}
\caption{Texting pedestrian and smartphone position.}
\label{fig:texting}
\end{subfigure} \quad \quad 
\begin{subfigure}[t]{.25\linewidth}
\includegraphics[height = 0.9\linewidth, width=0.8\linewidth]{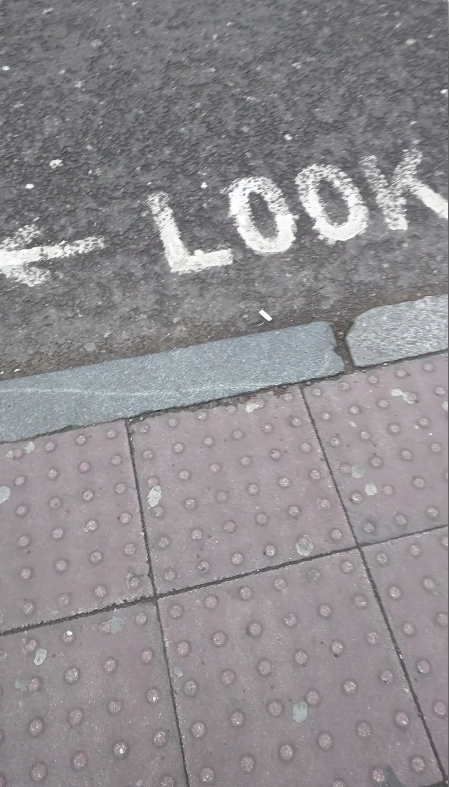}
\caption{Street sign captured during data collection in London.}\label{fig:camview}
\end{subfigure}
\caption{Smartphone camera field of view during texting.}
\label{fig:intro}
\end{figure}

Our focus is on studying whether texture recognition is possible on mobile phones using a pedestrian safety application as a case study. We are enabling smartphone and other mobile cameras to distinguish between materials that comprise real world outdoor walking surface, particularly streets and sidewalks.
A principal distinguishing factor between street and sidewalk is the material they are made of. One may therefore be tempted to attempt distinguishing street and sidewalk surfaces based on color. This approach is fragile, however, since the color perceived by a camera can change significantly depending on lighting conditions. We thus explore a more permanent characteristic, the \textit{texture} of the surface. When a pedestrian is texting, the rear camera is favorably directed to the ground in front of the user, thereby gazing at what the user may be walking on next, as in Figure~\ref{fig:texting}. We envisage using this camera opportunistically to identify the pedestrian's walking path, and distinguish between safe and unsafe walking locations. For all practical purposes, streets are considered unsafe for pedestrians, and sidewalks, safe. 

In this paper, we explore texture-based material recognition for mobile cameras. The primary challenge lies in enabling day-to-day smartphone cameras to distinguish between materials, such as that of sidewalks and street. We investigate the efficacy of light-weight texture descriptors on real world images, to leverage the subtle textural differences between materials that comprise real world streets and sidewalks. This encompasses a large set of images captured in various illumination and weather conditions. Also, the images are captured opportunistically, while the user is using the phone. We also consider ways of optimizing camera use without affecting performance, to conserve battery. 
We note that this paper focuses purely on the technical feasibility of the sensing aspect. Its interaction mechanism with the user and evaluation of user response will be left for future work.

Specifically, TerraFirma makes the following key contributions:

\begin{itemize}

\item demonstrating, through design and implementation, the feasibility of texture analysis for material classification on images captured using mobile cameras. 

\item developing a detection algorithm that can leverage textural features of the terrain to distinguish between street and sidewalk surfaces, and perform crossing detection.

\item gathering a first of its kind, unique database of street/sidewalk imagery across various countries. We collected camera footage in New York, London, Paris, and Pittsburgh. The entire dataset includes 10.5 hours of walking.

\item evaluating the performance of the proposed camera sensing system in various crowded high-clutter urban environments and comparing the performance of camera-sensing with dedicated inertial sensors on the same dataset.

\end{itemize}

%% file: background.tex
\section{Background and related work}

In this section we discuss the related work, and provide the necessary background for a gradient profiling technique. We compare how TerraFirma compares to this scheme. 

\input{related}

\subsection{Gradient Profiling for pedestrian safety}

A previous work, LookUp~\cite{LookUp}, addresses the challenge of detecting sidewalk-street transitions through a robust shoe-based step and terrain gradient profiling technique. Wearable sensing has penetrated the fitness tracking market. Shoe mounted sensors have been widely used for exercise tracking, posture analysis, and step counting~\cite{nikeplus,adidas}. LookUp uses similar sensors for sensing properties of the ground, and constructs ground profiles. 
Roadway features such as ramps and curbs~\cite{design} separate street and sidewalk, and hence detecting the presence of these features can help identify transitions from sidewalk to street and vice versa. Often ramps are present at dedicated crossings. These features are designed such that visually impaired pedestrians can distinguish sidewalks and streets.

LookUp leverages these roadway features to develop a sensing system that can automatically detect transitions from a sidewalk into the road~\cite{lookup_video, lookup_phdforum}. Importantly, it can track the inclination of the ground and detect the sloped transitions (ramps) that are installed at many dedicated crossing to improve accessibility. 


One of the advantages we have over LookUp is that the prototype requires additional hardware that includes a sensor unit to be mounted on the shoes.
LookUp acquires inertial data from the sensor, to detect changes in step pattern and ground patterns caused by ramps and curbs. In particular, a salient feature of this work is that it senses small changes in the inclination of the ground, which are expected due to ramps and the sideways slope of roadways to facilitate water runoff. 
Inertial sensors allow one to infer information with a very modest power budget, compared to GPS. The shoe-mounted sensor has the capability to measure the foot inclination at any given time. The inclination of the foot when it is flat on the ground is thus, the inclination of the ground.
Inertial sensor modules are mounted on both shoes, and share their measurements with a smartphone over a Bluetooth connection. Sensors on both feet substantially improve the detection of stepping over a curb, irrespective of the foot the pedestrian uses for the action. Although, ramp detections can be achieved even with a sensor on one foot. A smartphone serves as the hub for processing the shoe sensor data and implementing applications. 

LookUp processes raw accelerometer and gyroscope readings (sampled at 50 Hz) through a complementary filter, and extracts traces of pitch, yaw, and acceleration magnitude features from these measurements. While it primarily relies on this inertial data, it also collects magnetometer readings to assist with the Guard Zone filtering step.
Further, the pitch traces are divided into distinct steps and for each step cycle the period when the foot is flat on the ground, known as the stance phase, is extracted. The inclination of the foot during the stance phase represents the slope of the ground. Therefore, the slope of the ground is measured with every step of the pedestrian from the pitch readings.
The relative rotation of the foot is given by the yaw readings, during the stance phase. LookUp also extracts peak acceleration magnitude over an entire step cycle as an indication of foot impact force. Finally, it aims to detect stepping into the roadway through ramp and curb detection. Ramps are identified through characteristic changes in slope, while steps over a curb usually show higher foot impact forces. These candidate detections are then filtered through a guard zone mechanism. This mechanism helps to remove spurious events caused by uneven road surfaces. 

LookUp achieves higher then 90\% detection rates in the intricate Manhattan environment.
It  is  a  robust  solution to  sensing  sidewalk-street  transitions,  but  in  addition  to requiring  added  hardware,  it  also  does  not  perform  well when the sidewalk and street are at the same height.

%% file: related.tex
\subsection{Related Work}


Visual attributes have raised significant interest in the computer vision community. Specifically, texture-based material recognition have been advanced over the years~\cite{curet,amadasun} to the most recently proposed texture descriptors~\cite{texturesInTheWild, textons, deepfilterbanks}. Given the challenging nature of texture recognition, early work often focused on images captured using specific camera setups~\cite{curet}. Numerous datasets have been created that contain a diverse array of textures and materials. The more recent ones have been collected using images from the Internet~\cite{uiuc, drexel, texturesInTheWild, fmd}.

In earlier work on pedestrian safety, we have proven GPS~\cite{Jain_mars} and inertial sensors~\cite{ped_hotPlanet} on the smartphone to be insufficient for pedestrian safety in dense urban environments. Our wearable sensing approach LookUp!~\cite{LookUp} profiles the ground and detects street entrances via ramps and curbs. Although, it requires dedicated inertial sensors on shoes. 

Among camera-based approaches to pedestrian safety, obstacle detection based on smartphone camera, for the visually impaired~\cite{obstacle_visuallyImpaired} and for distracted pedestrians\cite{spareeye} is also common. These systems primarily work indoor to detect the presence of an object in the user path, and cannot classify  the path in front of the user. Crosswatch\cite{crosswatch} provides guidance to the visually impaired at traffic intersections. However, it requires the user to precisely align the camera to the crosswalk. 
Surround-see~\cite{surroundSee} is a smartphone system equipped with an omni-directional camera that enables peripheral vision.  
Smartphone cameras have also been proposed for use in indoor navigation~\cite{opticalFlowNavigation}. 
As discussed earlier, approaches such as Walksafe~\cite{walksafe} detects approaching vehicles when the pedestrian is already in-street, and only those approaching from one side. TypeNWalk~\cite{typenwalk} requires the user to watch the surroundings through smartphone camera while using an application, which already distracted users may not pay attention to. In addition, this approach requires the camera to be always on, thus increasing the battery consumption of the phone. The approach proposed by Tang et al~\cite{tactilepaving} is close to ours, but very different in that it assumes the presence of tactile paving at sidewalk-street transitions. As observed in our dataset collected across various cities, such tactile paving is not common.  

Texture based approaches are vogue in the fields of medical imaging, for example for detection of Diabetic Retinopathy~\cite{drPervasive}, an eye condition. Computer vision based texture analysis techniques are also widely used for defect analysis in civil infrastructure, such as concrete bridges and asphalt pavements~\cite{crackDetection_dana, concreteAsphalt}. 
These systems use digital cameras and dedicated mounting scheme for image capturing. 
Texture and other features on smartphones have been used for biometric recognition, such as iris scanning~\cite{iris}, food recognition~\cite{foodcam}. However, they require the user to directly point the camera and in some cases it also needs direct inputs from the user, such as bounding boxes~\cite{foodcam}.
Our proposed camera sensing system works opportunistically, with no user intervention, and amidst practical constraints such as camera motion, lighting changes, and blurring.

%% file: challenges.tex
\begin{figure*}[!t]
    \centering
     \begin{subfigure}[b]{.15\linewidth}
        \includegraphics[width=\textwidth]{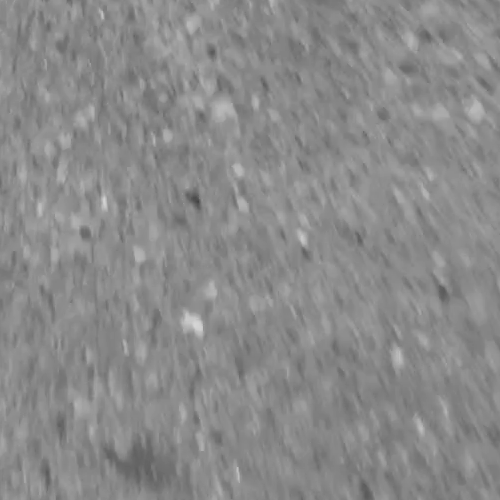}
        \caption{Asphalt}
        \label{fig:asphalt}
    \end{subfigure} \enskip
    \begin{subfigure}[b]{.15\linewidth}
        \includegraphics[width=\textwidth]{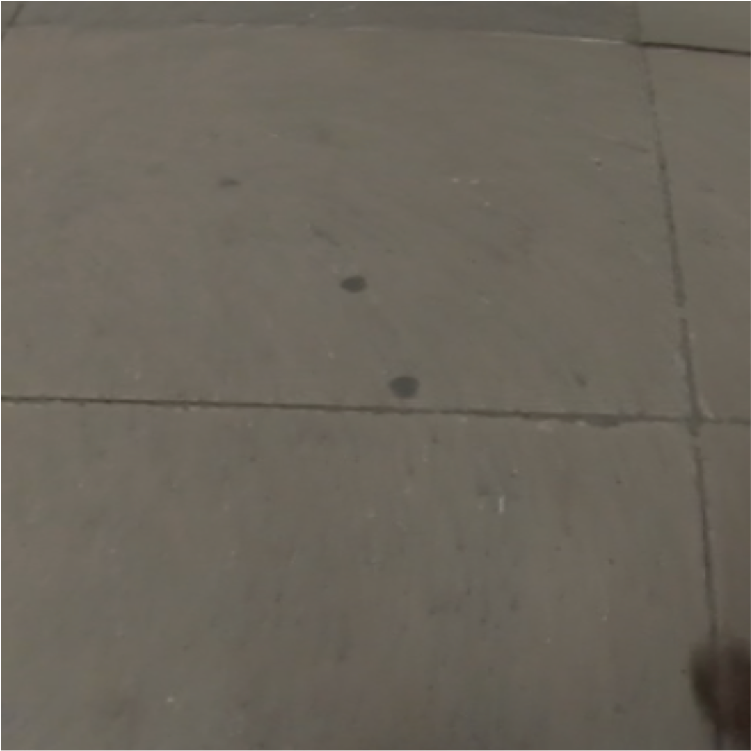}
        \caption{Concrete}
        \label{fig:brick}
    \end{subfigure} \enskip
    \begin{subfigure}[b]{.15\linewidth}
        \includegraphics[width=\textwidth]{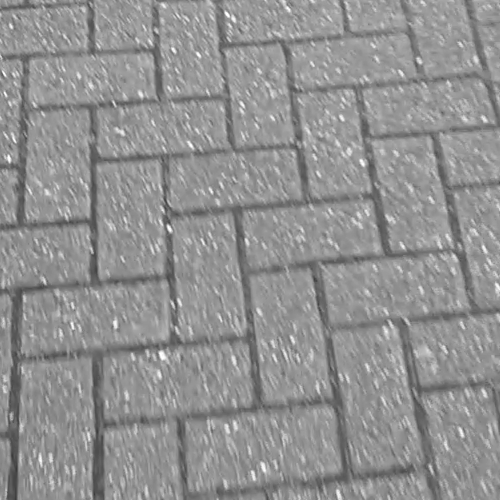}
        \caption{Brick}
        \label{fig:concrete}
    \end{subfigure} \enskip
    \begin{subfigure}[b]{.15\linewidth}
        \includegraphics[width=\textwidth]{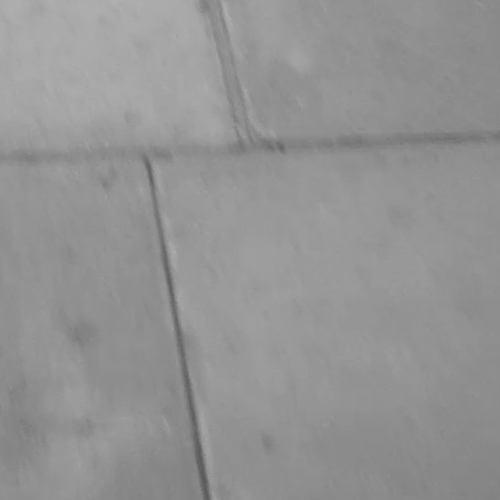}
        \caption{Tiles}
        \label{fig:tiles}
    \end{subfigure} \enskip
         \begin{subfigure}[b]{.15\linewidth}
        \includegraphics[width=\textwidth]{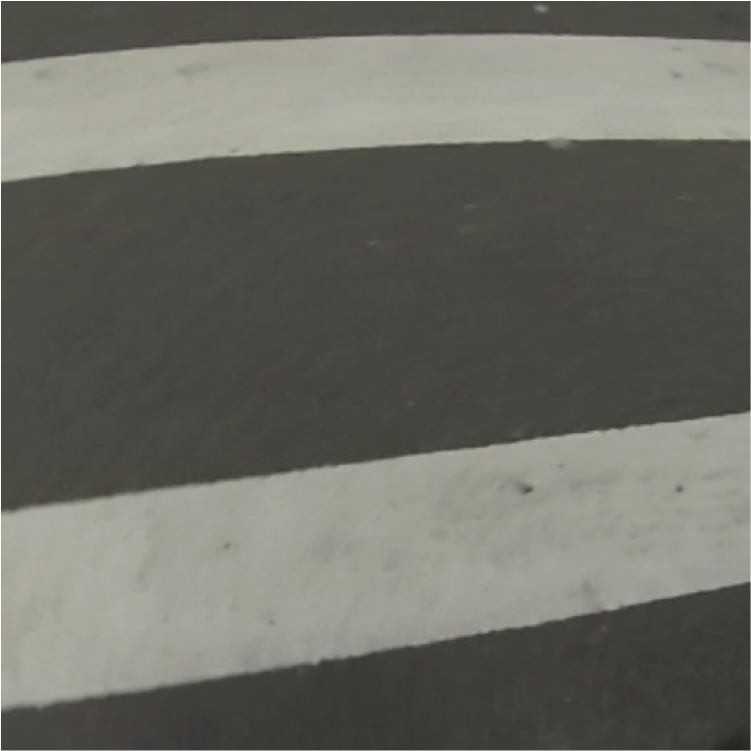}
        \caption{Ideal Crosswalk}
        \label{fig:crosswalk}
    \end{subfigure} \enskip
     \begin{subfigure}[b]{.15\linewidth}
        \includegraphics[width=\textwidth]{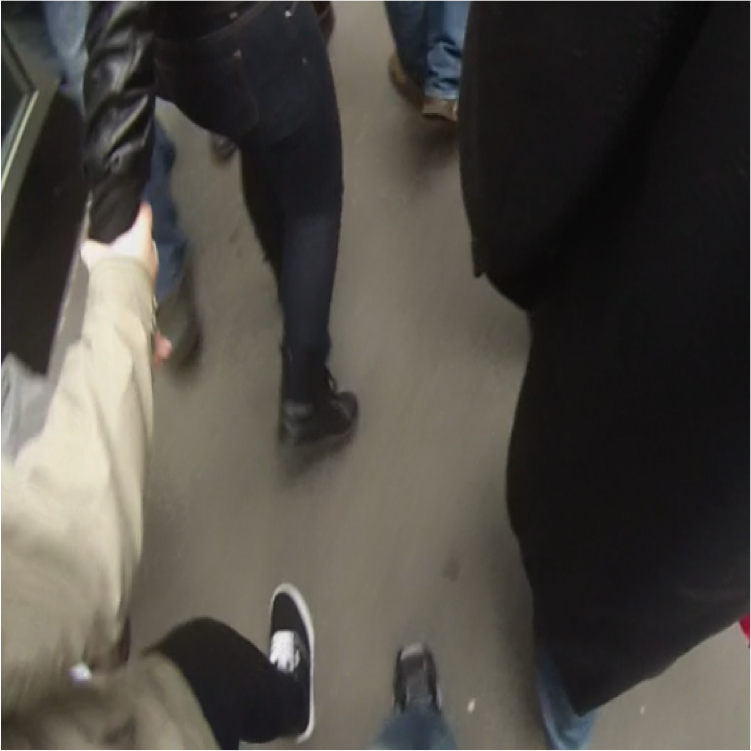}
        \caption{Crowded}
        \label{fig:crowded}
    \end{subfigure} 
    \par\medskip
    \begin{subfigure}[b]{.15\linewidth}
        \includegraphics[width=\textwidth]{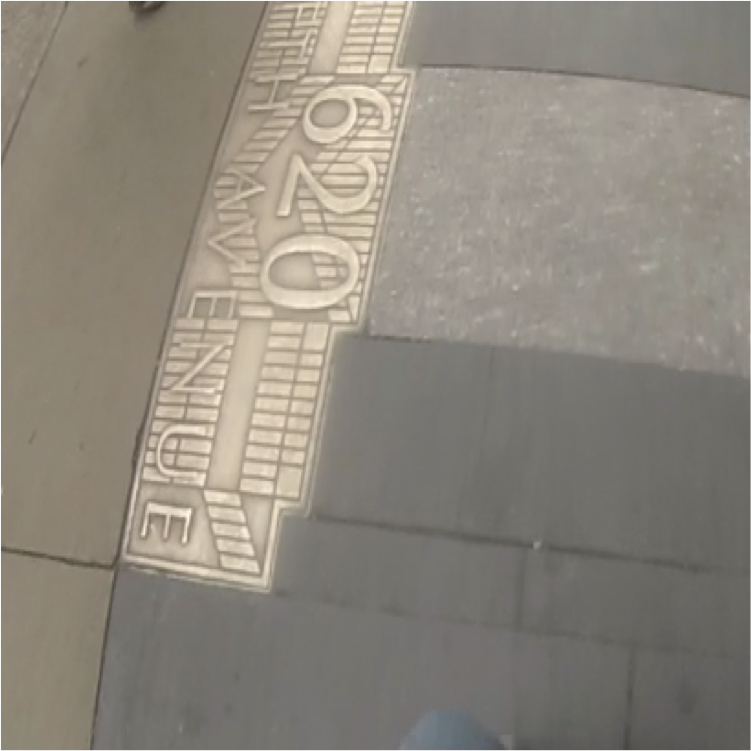}
        \caption{Diverse}
        \label{fig:diverse}
    \end{subfigure}\enskip
    \begin{subfigure}[b]{.15\linewidth}
        \includegraphics[width=\textwidth]{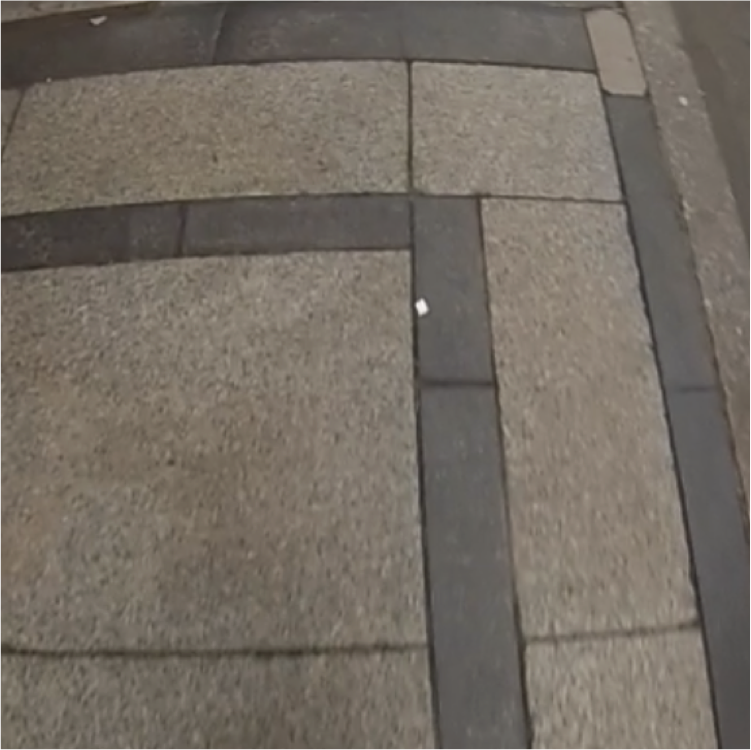}
        \caption{Tiled Pattern}
        \label{fig:pattern}
    \end{subfigure}\enskip
         \begin{subfigure}[b]{.15\linewidth}
        \includegraphics[width=\textwidth, height=\textwidth]{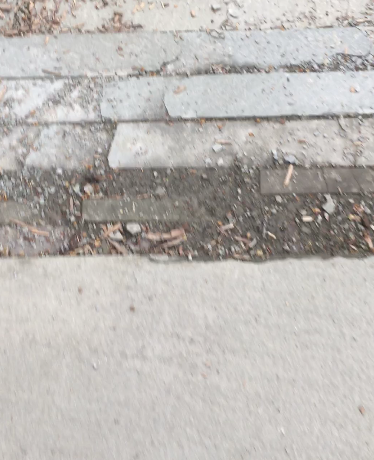}
        \caption{Broken sidewalk}
        \label{fig:crosswalk}
    \end{subfigure} \enskip
    \begin{subfigure}[b]{.15\linewidth}
        \includegraphics[width=\textwidth]{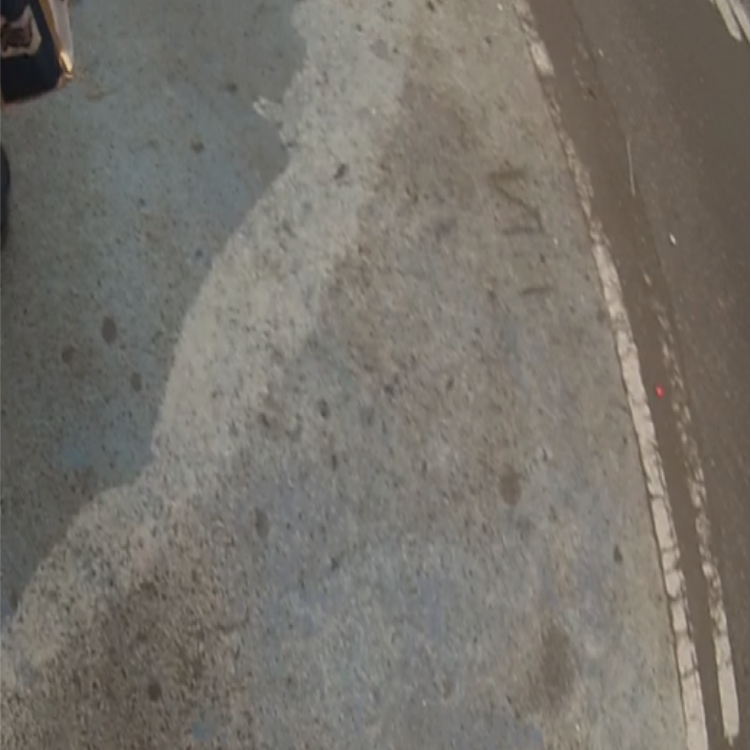}
        \caption{Defaced Street}
        \label{fig:defaced}
    \end{subfigure} \enskip
    \begin{subfigure}[b]{.15\linewidth}
        \includegraphics[width=\textwidth]{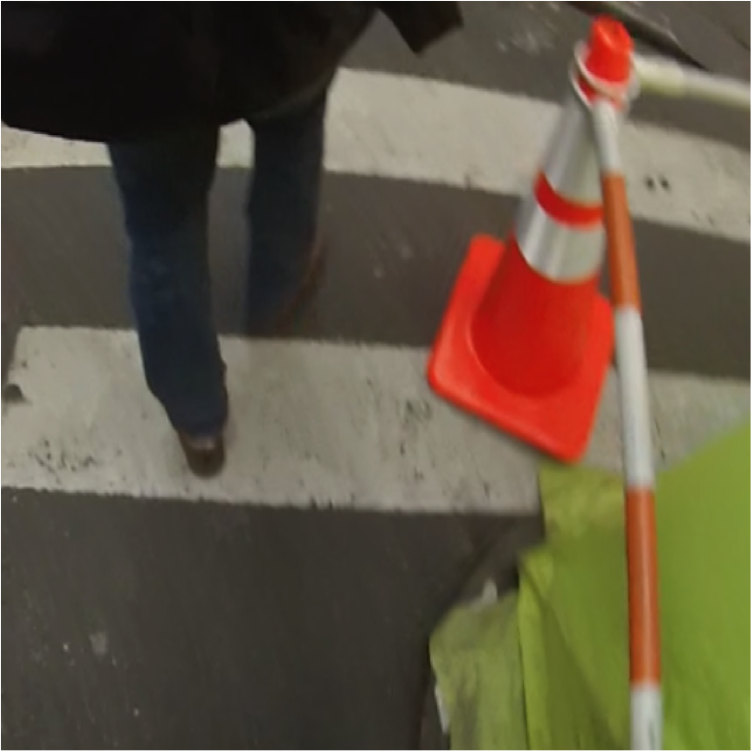}
        \caption{Cluttered}
        \label{fig:cluttered}
    \end{subfigure} \enskip
    \begin{subfigure}[b]{.15\linewidth}
        \includegraphics[width=\textwidth, height=\textwidth]{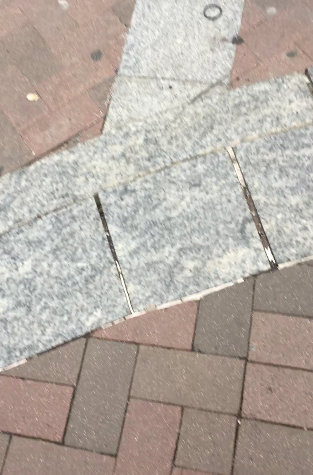}
        \caption{Same material}
        \label{fig:patched}
    \end{subfigure}
     \caption{(a)-(d): Samples of material classes found in our dataset. (e)-(l): Test examples from our dataset.}\label{fig:sidewalk}
\end{figure*}

\section{Applications}

Texture sensing through mobile cameras can benefit numerous applications. 

\textbf{Warning distracted pedestrians.} Sensing the texture of the ground that a person is walking on can be indicative of potential risk and can help mitigate it. Inattentive pedestrians can be alerted to watch out and be cautious as they transition from sidewalk to street without realizing. Opportunistically sensing such safety metrics also makes the application unobtrusive. 

\textbf{Pedestrian-Driver cognizance.} Identifying whether a pedestrian is on the sidewalk or in-street can vastly reduce the number of safety messages exchanged between pedestrians and drivers. When a pedestrian enters the street, this information can be communicated to an approaching vehicle. This helps in devising congestion control techniques, specially in dense urban areas, on wireless channels, such as DSRC.

\textbf{Infrastructure health monitoring.} Camera sensing on the smartphone can be used to crowdsource comparable information about the condition of the sidewalks and streets. Current practices require an official to manually inspect the health of these structures, which is time-consuming. An automated system that employs smartphone cameras can warn the city planning authorities when they detect perilous artifacts, such as potholes, bumps, hindrances to wheelchair accessibility, and absence of street lamps.

\textbf{Pedestrian localization through landmarks.} Large scale crowdsourced data can be used for enhancing outdoor localization by creating a material-based street/sidewalk map of the entire city. This system can be complementary to existing GPS based positioning.   

\textbf{Enhancing vision-based navigation.} Vision-based systems are increasingly dominating the autonomous guidance and navigation arena, ranging from cars~\cite{ny_selfdriving, onecam} to visually impaired pedestrians~\cite{crosswatch}. Understanding mobile camera based material recognition can augment these systems based on precise knowledge of which materials are easier to recognize, depending on changing light conditions. Similar analysis can be used for the visually impaired to learn which surfaces are more identifiable, particularly in bad weather, or cluttered environments. 


\section{Challenges}

Texture based recognition renders to be a harder problem than object recognition, because the real world ground surfaces often do not have a well defined shape, or predictable markers such as edges and corners. 
In distinguishing between materials that streets and sidewalk are made of, and to detect transitions from sidewalk to street, we encounter the following vital challenges. 

\textbf{Lack of standard paving materials.} 
Streets and sidewalks are not always constructed with the same type of material. In the United States, concrete slabs are more common for sidewalk constructions, however, we encountered stretches put together with tiles of different material, color, size and shape. We also observed sidewalks originally made of concrete slabs that were patched with asphalt in many locations. The lack of a standard guideline for which materials must be used in paving sidewalks and streets, and frequent changes in local policies~\cite{princeton} leaves our city sidewalks and streets full of diverse materials.


\textbf{Crosswalks as extension of sidewalk.} At many designated crossings, crosswalks are constructed with the same material as the sidewalk. For example, in Pittsburgh and London, crosswalks are also often made with bricks when sidewalks are made of bricks. In such cases detecting the material alone is not sufficient to identify when a pedestrian transitions from sidewalk to street.

\textbf{Sensitivity to Camera Motion.} Texture descriptors are popularly designed for sharp focused images taken from still cameras, and do not cope well with blurriness caused by motion. Since we aim to develop a functional technique for mobile cameras, blurriness in captured frames is a significant challenge.

\textbf{Environment noise.} Street and sidewalk appearances are impacted by lighting conditions and shadows. 
The visibility of the surface is immensely affected by the time of the day, presence or absence of street lights and the reflection of these lights from the pavement or street. In addition to just the variations in the surface appearance, there may be numerous conflicting objects in the camera view. A few noisy samples from Manhattan dataset are shown in Figure~\ref{fig:crowded} and Figure~\ref{fig:cluttered}.



\textbf{Energy consumption of the camera.} Camera based approaches commonly suffer from their high energy demand. Although the system should capture images only when the pedestrian is walking outdoor and actively using the phone, continuous use of camera can severely affect the battery life, and must be optimized in the best possible way.

%% file: method.tex
\section{System Design}



\begin{figure*}[t]
    \centering
        \includegraphics[scale=0.6]{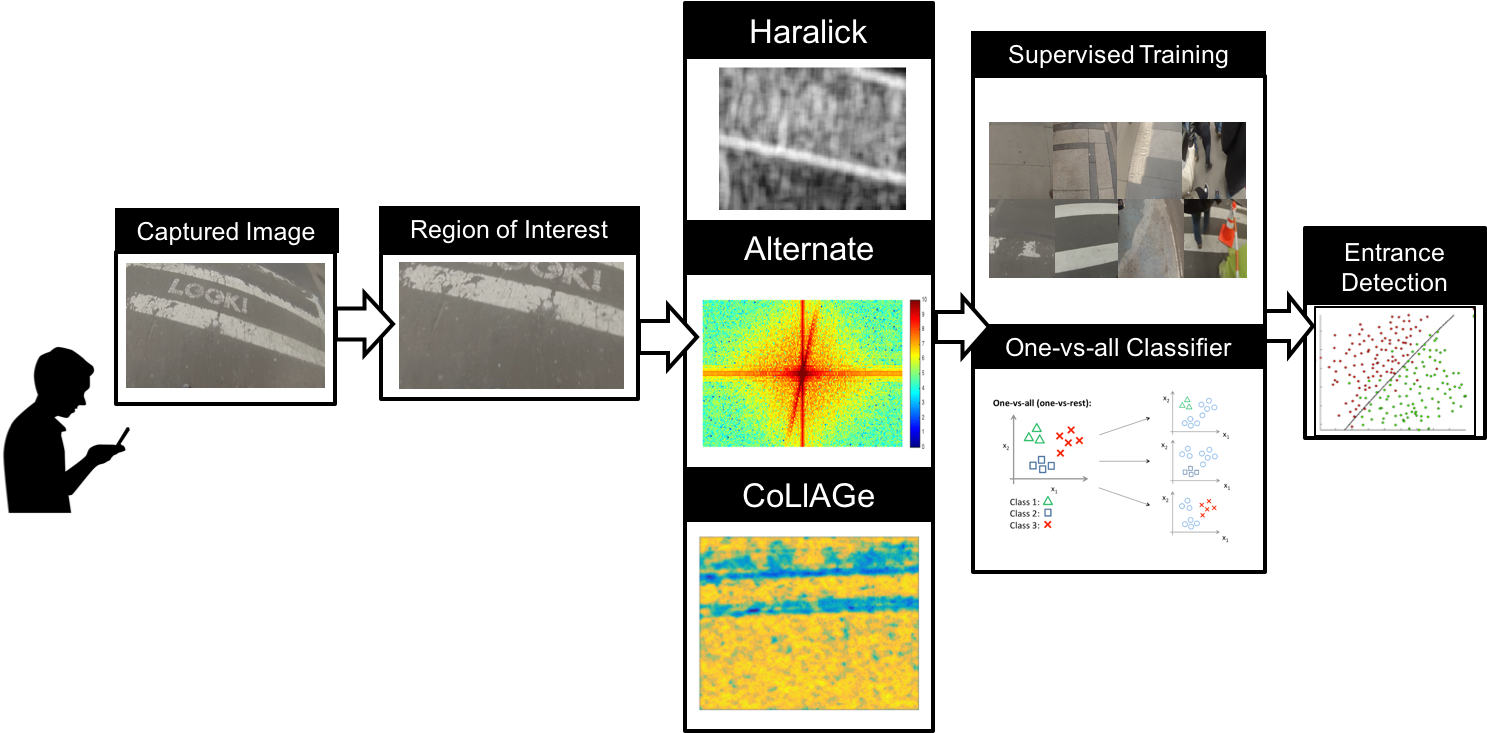}
        \caption{System Overview.}
        \label{fig:overview}
\end{figure*}

We propose a mobile camera based approach that aims to overcome the aforementioned challenges by deducing subtle textural features of the walking surface, to distinguish between paving materials. To achieve this, we leverage the smartphone position when a pedestrian is texting and walking, to opportunistically capture images of the ground ahead. Instead of deploying  object recognition techniques, we remove the dependence on the presence of external objects in the pedestrian's path, such as lamp posts, or tactile paving on ramps.
Figure~\ref{fig:overview} displays the flow of our system and the major steps involved. The camera on a distracted walker's smartphone captures snapshots. We capture snapshots instead of a video to conserve the smartphone battery. This task must be carried out in the background, without influencing how the user is interacting with the smartphone. This action can be triggered by sensing that the user is outdoor, in motion, and is actively using the smartphone. Such information can be gathered by minimally using the GPS, inertial sensors and checking if the screen is on, respectively.

The captured image contains a view of the path that lies in front of the user. This image may also capture objects in its view, such as trash cans or even the people walking around. A few examples of such images are shown in Figure~\ref{fig:sidewalk}.  
We extract a fixed sized patch, $\mathcal{R}$ around the center of the image, since this region is most likely to be free of environment clutter. 
Further, we compute visual information from $\mathcal{R}$, known as features. These features identify the type of surface the camera is looking at, or in other words, the \textit{texture}. We borrow our feature computation techniques from the computer vision community~\cite{glcm_texture, collage}. This texture information can be used to distinguish between the paving materials found across our cities. In addition to distinguishing these materials, we also aim to identify when a pedestrian transitions from sidewalk to street.
While it is easy for the human eye to distinguish streets and sidewalks, it is rather challenging to train a camera to acknowledge the differences. To account for the subtleties in these features and to ensure robustness, we use a supervised learning algorithm, that can classify the extracted image as one of the paving materials.
We discuss the details of our feature selection process and classification in the following section

\section {Understanding Texture}
\label{sec:features}
Texture is a visual attribute of images, where unlike objects, the overall shape is not important. Texture captures local intensity statistics within small neighborhoods. These can then be used to quantify the smoothness, or `feel' of the surface. Texture information adds a layer of detail beyond object recognition. 

\subsection{Patch Selection}

Often, captured frames include a view of the people walking around, lamp posts and garbage cans that line the sidewalks, and amount to clutter in the  image. They provide little or  no information about the ground surface ahead, and thus we preprocess our captured frames to get rid of clutter. This also helps us operate on more salient regions of the image. We observe that these objects usually border the perimeter of the image and the patch of land right in front of the pedestrian is clear for him to take the next step. Based on this observation, we attempt to reduce noise in each frame and get a clear view of the path by extracting a fixed size region, $\mathcal{R}$, around the center from each frame.
$\mathcal{R}$ is a square patch of size ${n}\times{n}$, and captures the paving of the ground ahead with little or no clutter.
We found a patch size of $500\times500$ around the center of the image to be optimal for capturing the scene ahead, and small enough so as not to cause significant computational overhead.
All further operations are conducted on the image $\mathcal{R}$.

\subsection{Texture Representations}

In this section we discuss the feature descriptors computed on the image region $\mathcal{R}$ obtained in the previous step.
We hypothesize that even though the gray level intensities are similar, the localized pixel-wise arrangement will be significantly different across the materials used for paving streets and sidewalks. We found three types of texture descriptors to be suitable for assessment:

\begin{figure}
    \centering
    \begin{subfigure}[b]{.45\linewidth}
        \includegraphics[width=\textwidth]{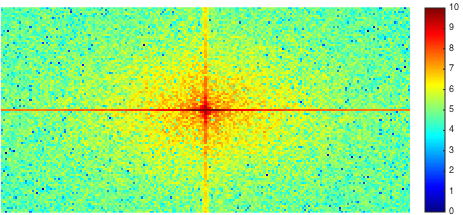}
        \caption{Street (No Crosswalk)}
        \label{fig:street_fft}
    \end{subfigure}
    \begin{subfigure}[b]{.45\linewidth}
        \includegraphics[width=\textwidth]{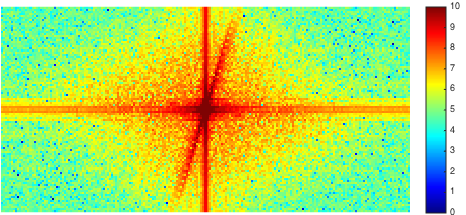}
        \caption{Crosswalk}
        \label{fig:crosswalk_fft}
    \end{subfigure}
    \begin{subfigure}[b]{.45\linewidth}
        \includegraphics[width=\textwidth]{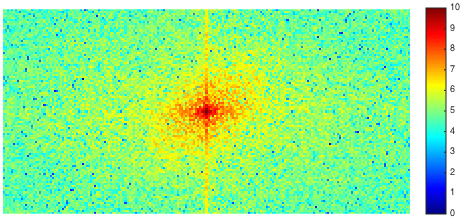}
        \caption{Concrete sidewalk}
        \label{fig:concretesw_fft}
    \end{subfigure}
     \begin{subfigure}[b]{.45\linewidth}
        \includegraphics[width=\textwidth]{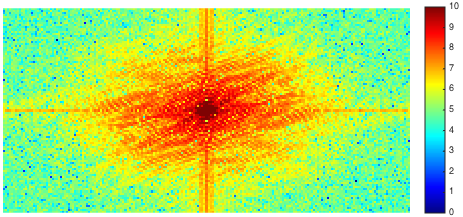}
        \caption{Crowded sidewalk}
        \label{fig:crowded_fft}
    \end{subfigure}
     \caption{Frequency Domain Representations. Samples from New York Dataset.}\label{fig:fft}
\end{figure}

{\bf Haralick:} Haralick~\cite{haralick} texture descriptors have proven to be robust across various datasets for texture recognition. Compared to recent texture descriptors, we were impressed with the performance of these low-level texture descriptors on our real world dataset. Gray-Level Co-occurrence Matrix features (GLCM), can be used to extract Haralick features. The GLCM is given by NxN matrix $M$, where N is the number of gray levels in the image. Each matrix element $p(i,j)$ is the probability of pixel with intensity value i being adjacent to that with intensity value j. We extract 13 Haralick descriptors~\cite{haralick} from $M$. Each of these captures a unique property of the image. These descriptors calculate the relationship between a reference pixel and its neighbors in a 2D matrix and second order statistics thereof.
For example, entropy feature is given by $\mathcal{E}(i,j) = -\sum_{i}\sum_{j} p(i,j) log( p(i,j))$.
$\mathcal{E}$ calculates the degree of disorder in the image. For an image where the transitions are very high, the corresponding $\mathcal{E}$ is also high and vice versa.
Each Haralick feature computation returns $n\times{n}$ values computed for $\mathcal{R}$, where $n$ is the number of pixels along each dimension. We encode these per-pixel local descriptors into global feature values by summing their values over all pixels. We obtain the Haralick feature vector $\mathcal{V_H}$, which is a 13-dimensional vector, for each image $\mathcal{R}$.

{\bf CoLlAGe:} Co-occurrence of Local Anisotropic Gradient Orientations (CoLlAGe)~\cite{collage, collage_nature}, a recently introduced texture descriptor in the field of biomedical imaging, has shown promise in distinguishing benign and malignant tumors from anatomic imaging. It captures higher order co-occurrence patterns of local gradient tensors at a pixel level. CoLlAGe is different than traditional texture descriptors in that it accounts for gradient orientations at a local scale, rather than at a global scale. Mathematically, CoLlAGe computes the degree of disorder in pixel-level gradient orientations in local patches. 
As in~\cite{collage}, for every pixel $c$, gradients along the $X$ and  $Y$  directions are computed as:

\begin{equation}
\nabla f(c)= \frac{\partial f(c)}{\partial X}\hat{i} + \frac{\partial f(c)}{\partial Y}\hat{j}
\end{equation}
where $ \frac{\partial f(c)}{\partial q}$ is the gradient magnitude along the $q$ axis, $q \in \{X,Y\} $. A ${N} \times {N}$ window centered around every $c \in C$ is selected to compute the localized gradient field. 
We obtain dominant principal components from the vector gradient matrix, to compute the principal orientation for each pixel. This captures the variability in orientations across $(X,Y)$. Then individually 13 Haralick statistics are computed as shown in ~\cite{haralick}. For every feature, first order statistics (i.e. mean, median, standard deviation, skewness, and kurtosis) are computed. Variance  is  a  measure  of the  histogram  width that measures the deviation of gray levels from the Mean. Skewness is a measure of the degree of histogram asymmetry around the Mean and Kurtosis is a measure of the histogram sharpness. Computing five statistics for each of the 13 features, gives us the CoLlAGe feature vector $\mathcal{V_C}$, which is a 65-dimensional vector.

\begin{figure}[!t]
    \centering
    \begin{subfigure}[b]{.45\linewidth}
        \includegraphics[width=\textwidth]{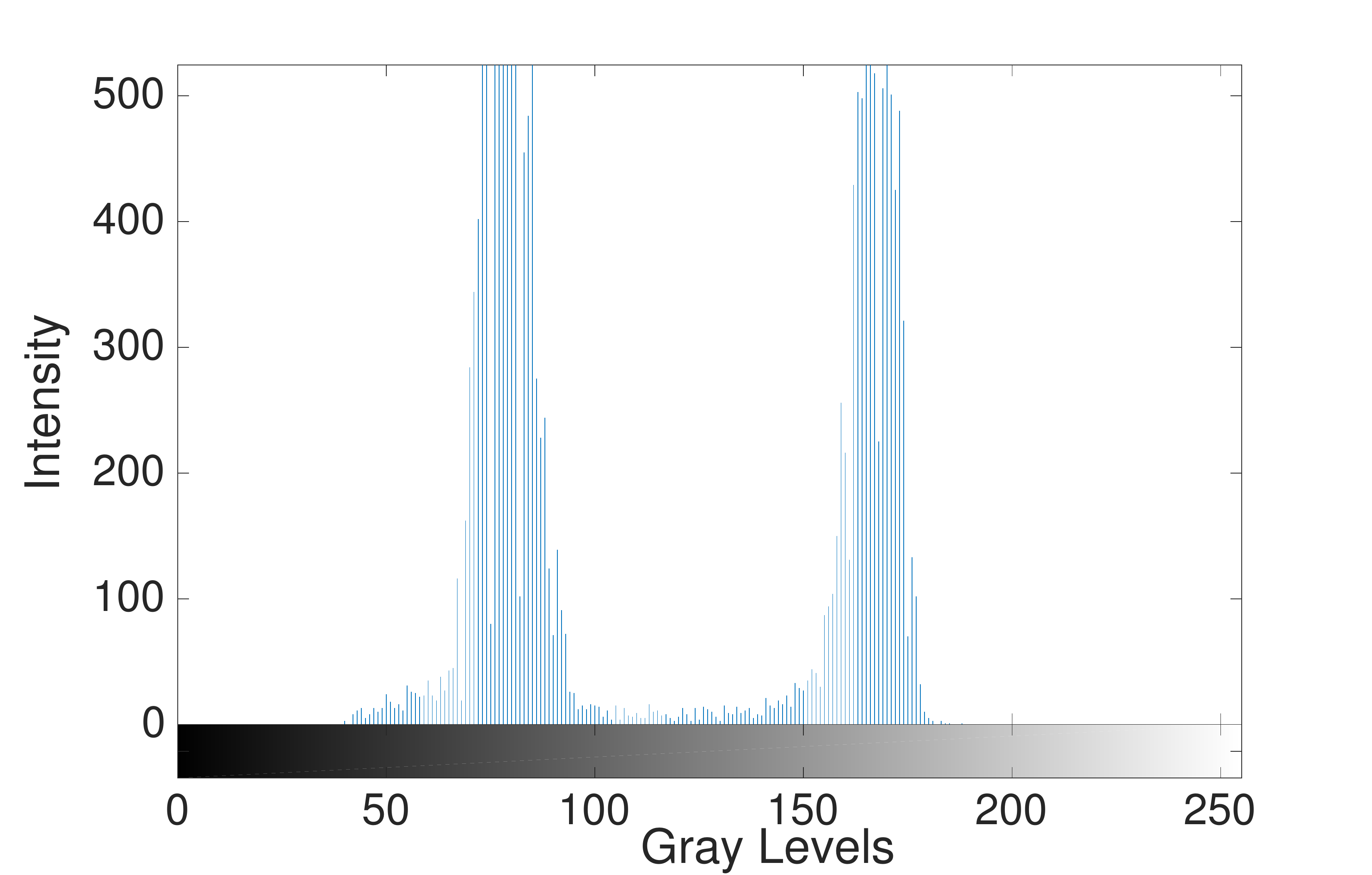}
        \caption{Crosswalk}
        \label{fig:streetHist}
    \end{subfigure}
    \begin{subfigure}[b]{.45\linewidth}
        \includegraphics[width=\textwidth]{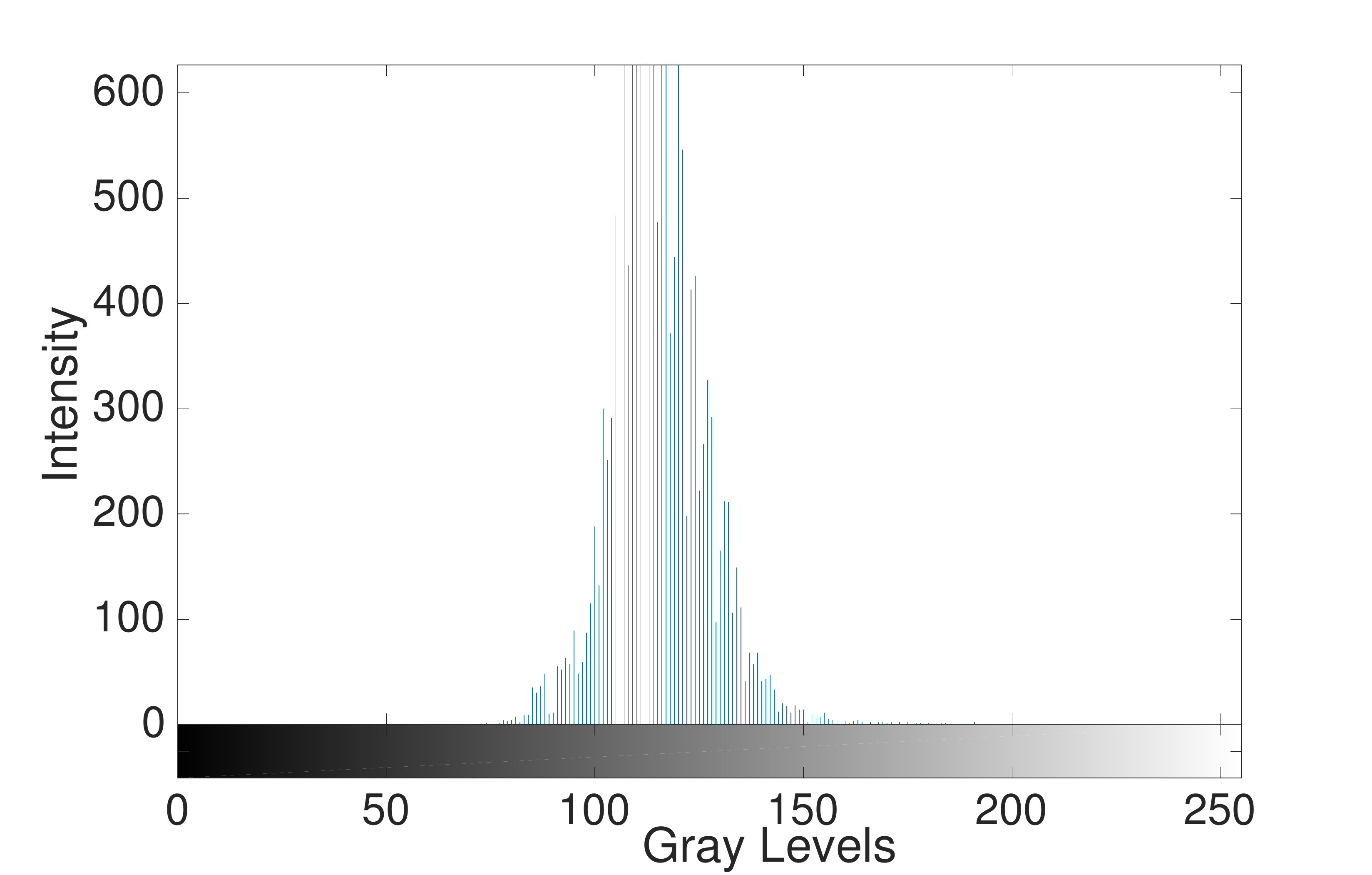}
        \caption{Concrete Sidewalk}
        \label{fig:sidewalkHist}
    \end{subfigure}
    \begin{subfigure}[b]{.45\linewidth}
        \includegraphics[width=\textwidth]{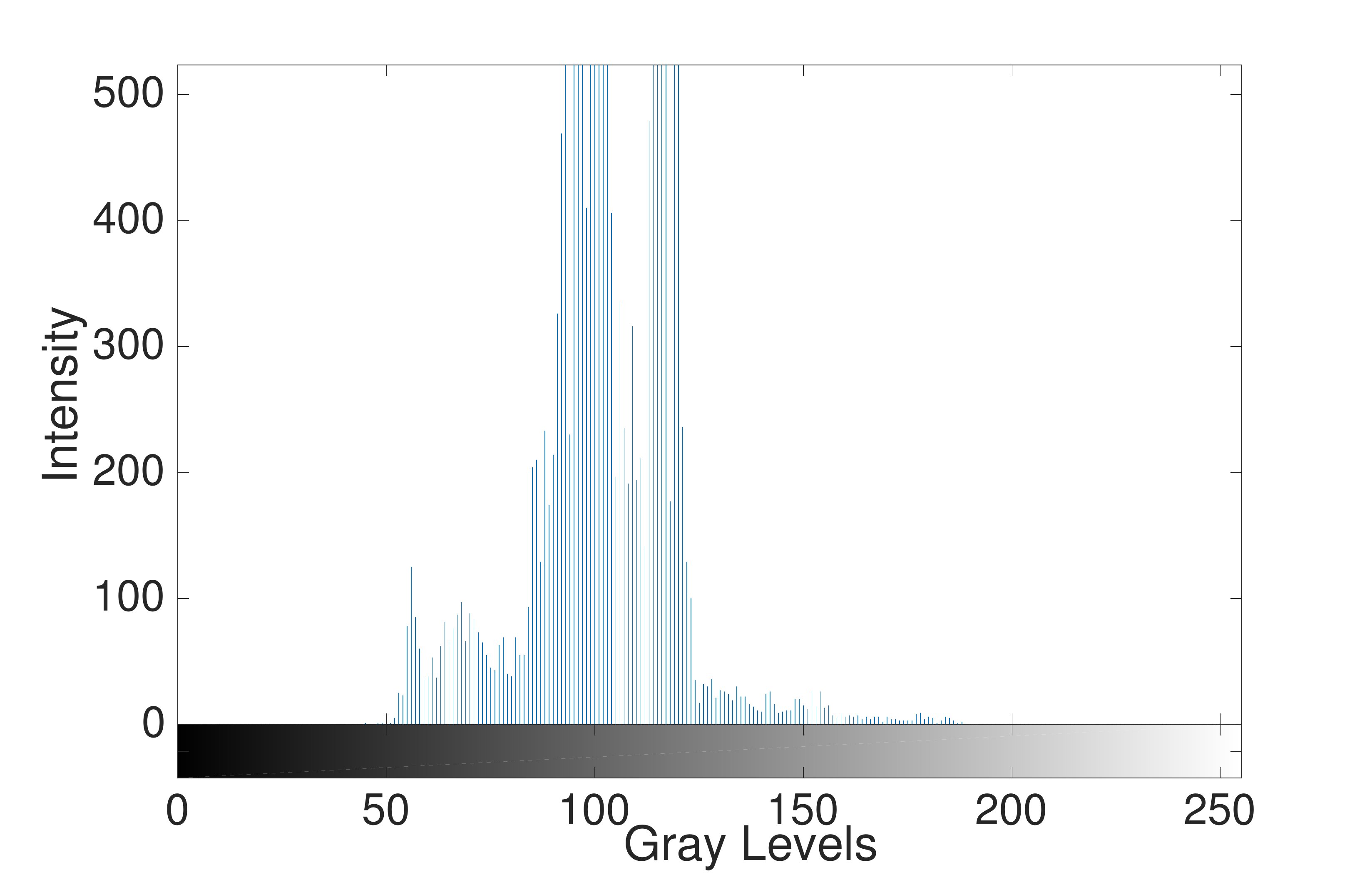}
        \caption{Crowded Sidewalk}
        \label{fig:crowdHist}
    \end{subfigure}
     \caption{Intensity Histograms.}\label{fig:hist}
\end{figure}

{\bf Alternate representations:} We observe that analyzing alternate representations of the image provides significant information about the image pattern. We consider the following values for each image in addition to the features mentioned above:

- {\it Prominent peaks in gray-level intensities:} The number of peaks in the intensity histogram, and the spacing between them can be used to distinguish a cluttered image from one with a pattern, possibly a crosswalk. To this effect, we calculate the number of peaks in the image intensity histogram, with a minimum height of 200 and at least 60 intensity levels apart. These thresholds were computed empirically. Sample intensity histograms from our New York City data are shown in Figure~\ref{fig:hist}.

- {\it Fourier domain features:} Fourier domain analysis provides us with the frequency domain representation of the image.
A 2D Fourier transform yields the frequency response image indicative of the magnitude and phase of the underlying frequency components. The Fourier transform $F$ of an $M$$\times$$N$ image is given by

\begin{equation}
F(p,q) = \frac{1}{MN}\sum_{i=0}^{M-1}\sum_{j=0}^{N-1} f(i,j) e^{{-j2{\pi}}{(\frac{pi}{M} +\frac{qj}{N})}}
\end{equation}

We take the magnitude image and compute first order statistics from it (mean, median and skewness) as our Fourier features. Fourier magnitude representative images are shown in Figure~\ref{fig:fft}. For the ease of visualization, the zero frequency component has been shifted to the center. Furthermore, the magnitudes of these frequency components have been normalized to 10 levels and color coded. It can be seen that a crowded sidewalk has a large number of low frequency components, compared to a plain concrete sidewalk. The crosswalk pattern is also identifiable and distinguishable from others, as seen in Figure~\ref{fig:crosswalk_fft}.

- {\it Range filter features:} Instead of looking at the absolute range of pixel intensities, we captured first order statistics like mean and median within a ${3}\times{3}$ neighborhood of all the pixels. 
After extracting features from alternate representations, we obtain the $16$-dimensional feature vector $\mathcal{V_A}$.

\subsection{Feature Selection and Classification}

\begin{figure*}[!t]
    \centering
    \begin{subfigure}[b]{.3\linewidth}
        \includegraphics[width=\textwidth]{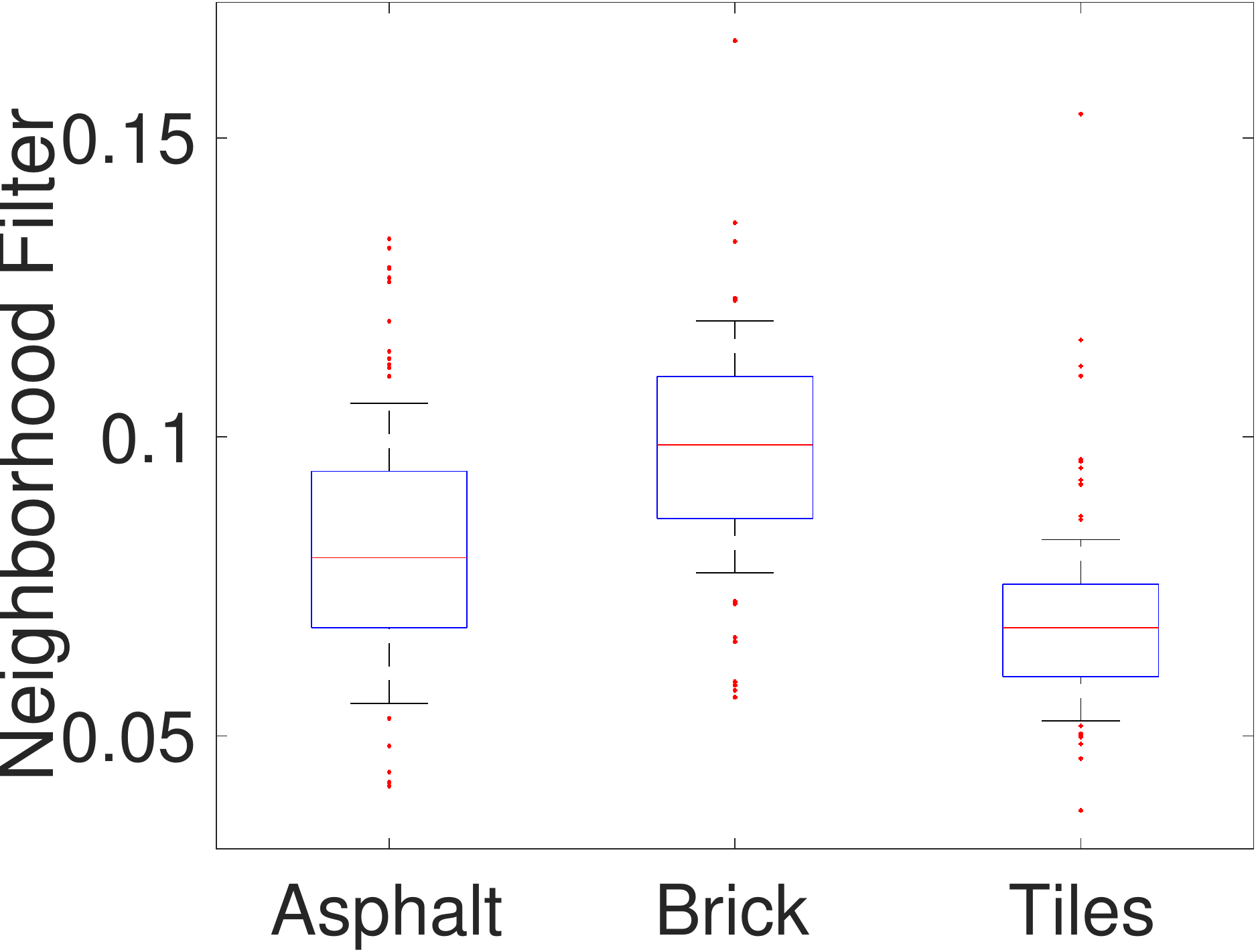}
        \caption{London}
        \label{fig:london_boxplots}
    \end{subfigure}\quad
    \begin{subfigure}[b]{.3\linewidth}
        \includegraphics[width=\textwidth]{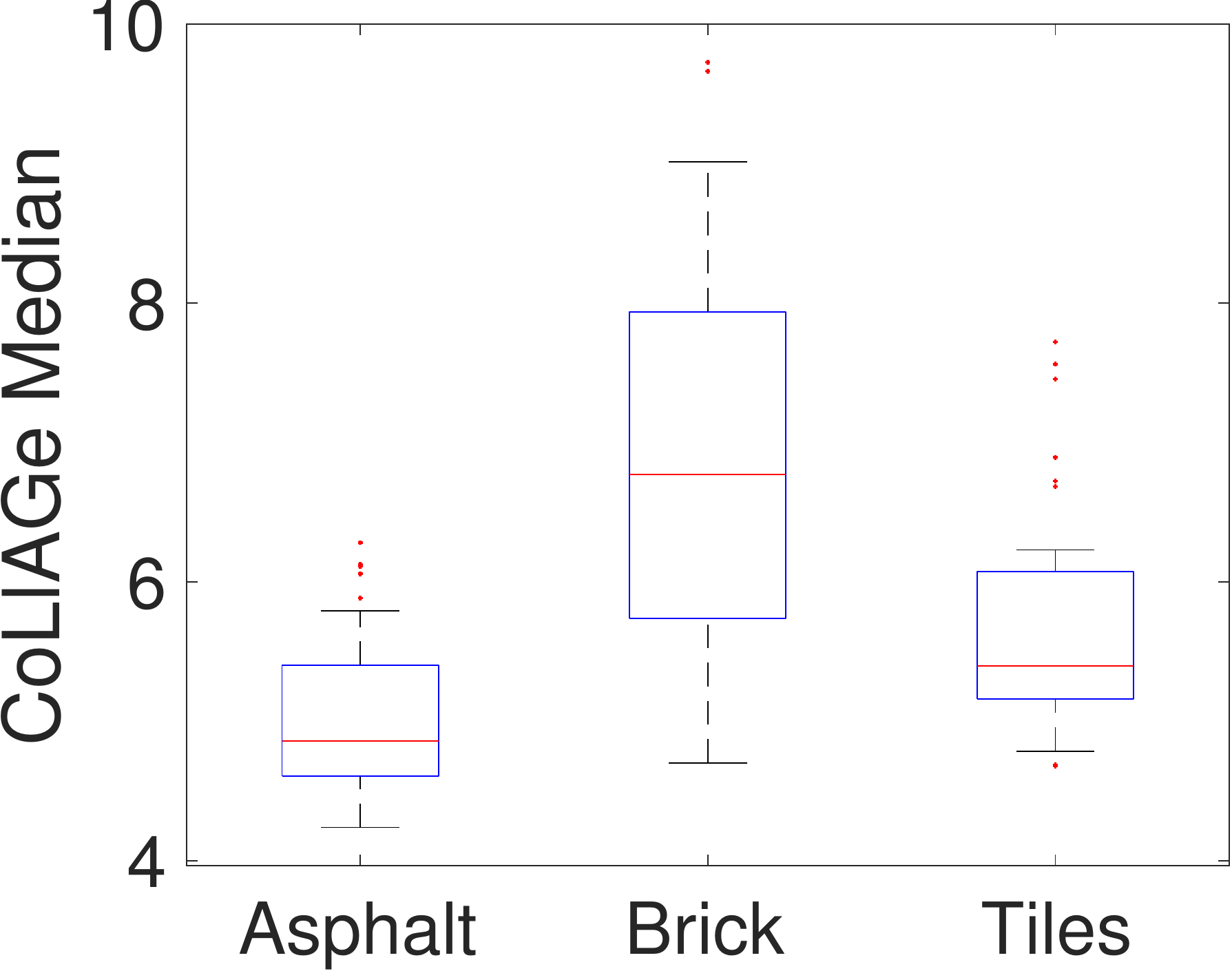}
        \caption{New York.}
        \label{fig:ny_boxplots}
    \end{subfigure}\quad
    \begin{subfigure}[b]{.3\linewidth}
        \includegraphics[width=\textwidth]{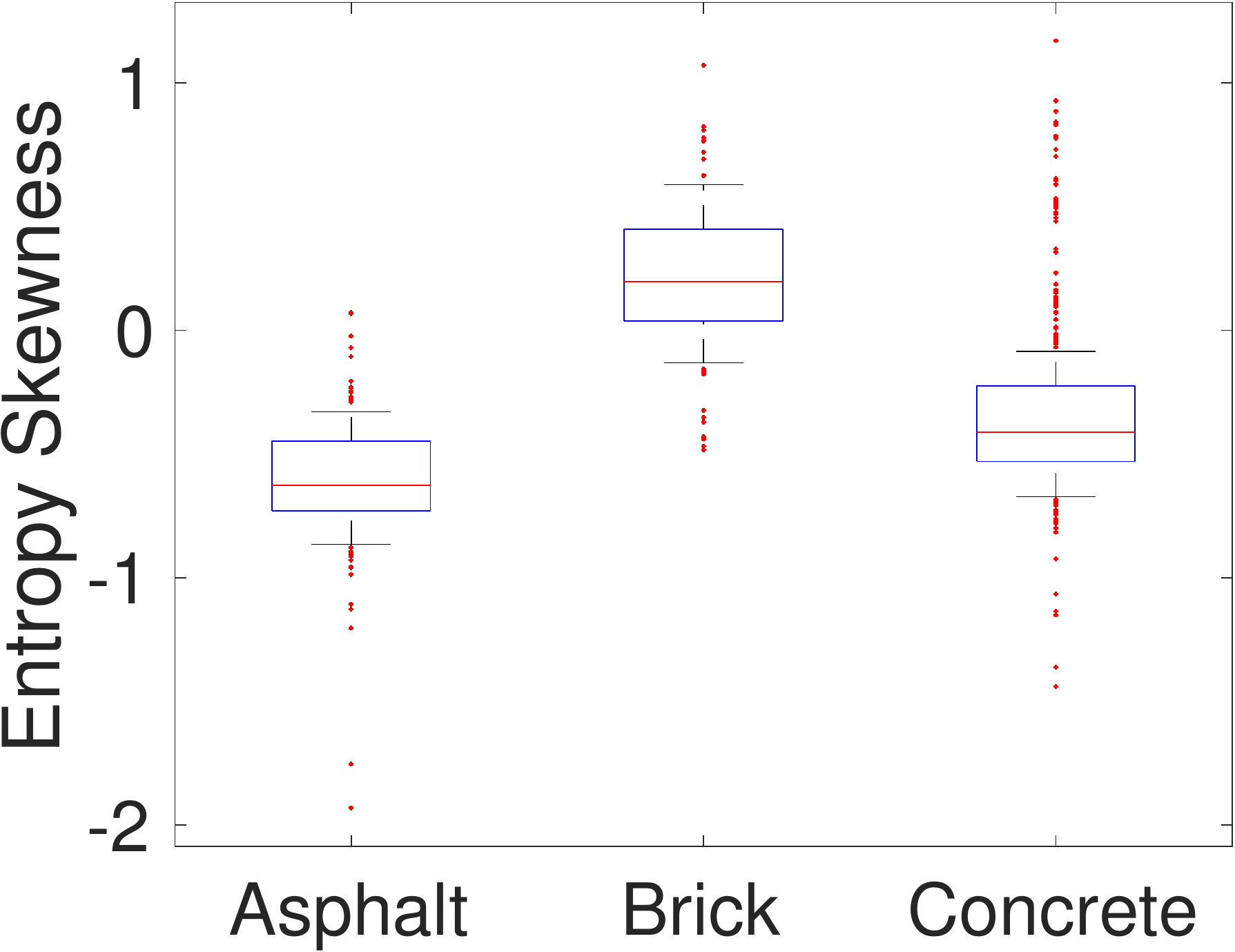}
        \caption{Pittsburgh.}
        \label{fig:pitts_boxplot}
    \end{subfigure}
     \caption{Feature selection to identify the most distinguishing feature in each dataset.}\label{fig:box}
\end{figure*}

For each captured frame $\mathcal{R}$, we compute the feature descriptors mentioned above, and aggregate them to form the feature space $\mathcal{FS} = \{\mathcal{V_A}, \mathcal{V_H}, \mathcal{V_C}\}$, with $94$ features.
To account for the `curse of dimensionality', the descriptors in  $\mathcal{FS}$ are dimensionality reduced by using Principal Component Analysis (PCA). This maps the high-dimensional feature space to a new space with reduced dimensionality, known as feature transformation. We quantitatively analyze our features to identify the most distinguishing features in each test environment, irrespective of the learning algorithm. We use the Wilcoxon rank sum test~\cite{ranksum} for each corresponding feature in each pair of classes, and select the  one with the lowest p-value. As we can see in figure~\ref{fig:box}(a), the most distinguishing feature among all materials found in London is the standard deviation of the range filter applied to the image. Similarly, we see in figure~\ref{fig:box}(b) that CoLlAGe median separates the New York City data well.

The materials are classified using an error-correcting output codes (ECOC)~\cite{ecoc} classification method, which reduces the classification to a set of binary learners. We used a one-vs-all coding design, and Support Vector Machine (SVM) classifiers with linear kernels as binary learners. Despite our biased dataset due to disproportionate occurrences of materials across cities, we train an unbiased classifier to avoid overfitting to any one class. We provide the classifier with equal number of samples from each class. We train a separate ECOC classifier for each test city. The model is validated using ten fold cross validation. Furthermore, we perform Platt sigmoid fitting~\cite{platt} on top of the SVM results, i.e. the scores returned from cross validation. We estimate the posterior probabilities by entering the cross-validation scores into the fitted sigmoid function. The results are reported in Section~\ref{section:eval}.

\subsection{Street Entrance Detection}

When pedestrians transition from sidewalk to street, the material of the ground may or may not change. We observed a few common scenarios. First, if the transition is made via a designated crossing, the street may have a crosswalk, which can be made of the street material (often asphalt), or may be made of the same material as the sidewalk (for example bricks in Pittsburgh). When made of asphalt, it may or may not  have alternate light and dark stripes (also called a zebra crossing). Second, when crossing via a curb, in most places there is a border that separates street or sidewalk. Even in our dataset from Paris, where most sidewalks and streets are both made of asphalt, a small concrete or tiled border separates sidewalks from street, as shown in Figure~\ref{fig:patched}. We aim to detect this transition border.
The frames captured during transition have multiple textures. To detect these frames with multiple textures, we create a small training set with just the transition frames, and train a one-vs-all classifier as described above, with transitions as the positive class and all other materials as negative class. We use this classifier on an unseen test sequence, and classify every $n^{th}$ frame as transition or not a transition. We find $n=6$, which is equal to $5 fps$ to give us optimal performance. Note that to conserve battery, we only capture images and do not record a video. Since many frames may have multiple textures, we get rid of false positives by using guard zone filtering, similar to that described in LookUp~\cite{LookUp}. When a transition is detected with a high score from the classifier, we reckon this to be a definite entrance into the street and mark it as a high confidence event. We set a guard zone following this detection, for $2$ seconds, which means that all detections within the next $2$ seconds are discarded. The guard zone value was chosen empirically based on our observation that a transition typically lasts $3-4$ seconds.

%% file: experiments.tex
\section{Dataset Design and Collection}

To evaluate the performance of our system, and to ensure robustness, volunteers from various metropolitan areas across the world to collected data while walking in their cities. The advantages of doing this were manifold. First, the data was collected by a diverse set of pedestrians, therefore accounting for variances in how pedestrians hold their phone, and walking behavior. Second, the data was collected on different models of smartphones, all of which had different camera specifications. Third, we were able to gather a large amount of data, leading to a unique database of street/sidewalk imagery from a pedestrian's perspective\footnote{Dataset available upon request.}. Fourth, due to the duration of our data collection efforts, our dataset comprises of dissimilar seasons, weather conditions, and illumination. Therefore, our test data was collected in the wild, on real smartphones, on people's daily walking paths, and not in controlled settings.

\begin{table}
\centering
 \begin{tabular}{|l | c |c | c | c|} 
 \hline
 {\bf  City} & {\bf Volunteers} &{\bf Duration} &  {\bf Sidewalk } &  {\bf Street } \\ [0.5ex] 
 \hline
 New York & 5 & 5h   & Concrete &  Asphalt \\ [0.3ex] 
 \hline
 \multirow {2}{*}{London} & 1 & \multirow {2}{*}{} 45m & Tiles & Asphalt\\
                              &  &    &  Brick & Brick \\[0.3ex] 
 \hline
 \multirow {2}{*}{Paris} & 2 & \multirow {2}{*}{}  3h 30m & \multirow {2}{*}{Asphalt} & Asphalt\\
 & & &  & Brick \\[0.3ex] 
 \hline
 \multirow {2}{*}{Pittsburgh} & 1 & 1h 10m  & Concrete & Asphalt\\
 & &  & Brick & Brick \\[0.3ex] 
 \hline
\end{tabular} 
  \caption{Pedestrian Dataset Summary.}\label{tab:dataset}
\end{table}

Table~\ref{tab:dataset} introduces details of our dataset, which is a collection of real-world videos recorded during the pedestrian's commute and daily chores. 
There is a wide variation in the angles that the phone was held in, which adds diversity to our test set. The videos capture the ground that the participant is walking on. In a typical texting position, the view of the smartphone's rear camera comprises the ground surface ahead of the user. To avoid any bias, participants were not made aware of the purpose of the data collection. They only captured videos during daytime, to ensure their safety during data collection.
They were asked to make recordings only when walking alone, to avoid capturing participants' conversations. However, as a precaution, all the collected videos were stripped of any audio before processing. 

The videos were manually labeled to annotate the exact frame number when (i) the pedestrian sets first foot into the street from the sidewalk- an entrance instance, and (ii) the pedestrian sets first step from the street to the sidewalk- an exit instance. The entrance instants were used as ground truth for correctness and timeliness of street entrance detections. 
The entrance and exit instances, together, were used to divide each video into street, sidewalk, and transitions. Further, streets and sidewalks were manually subdivided into the materials they were made of. Moreover, for training only those images were retained where the primary material fills at least 80\% of the image. This is to ensure that we can study texture recognition independent of texture segmentation. However, to maintain the characteristics of an in-the-wild dataset, we retain images with ground artifacts, for example, tree trunks, manhole covers, poles, ramp grates etc, but no crowds.
The frames retained as transitions, were those that covered the sidewalk-street transition. These were frames where more than one material was visible in the camera view. They capture the street sidewalk separators, such as ramps and curbs. All ramps, without or without tactile paving were captured. 


In New York City, the data was collected from the midtown area of Manhattan. This is the same data set as used in LookUp!~\cite{LookUp}. 
The camera sensing data was collected by five volunteers, who traversed a 2.1 miles long path. Half the data was collected during the day, while the other half was collected after sunset. The average time taken to complete each loop was about 60 mins and involved 32 crossings. The data was collected at various times, including weekday rush hours and weekends. For our experiments in this paper, we only used the data collected during daytime. This dataset was collected using a GoPro Hero 3 camera. This camera was placed upon the pedestrian, using a chest harness, and oriented to point downwards, simulating the texting position of a smartphone. 
The GoPro recorded video at 60 frames per second at a resolution of 720p. 
We subsample this high frame rate data for our analysis and evaluation, as presented in the next section.
In London, the data was recorded using a Nexus 6, during daily commutes and weekend  chores over a period of two months. 
In Pittsburgh, the data was recorded in one 70-minute long walking session using an iPhone 6s. It was recorded around dusk, in the downtown area, and also covers two bridges.
In Paris, the data was recorded using an LG Nexus 5 during the early morning and afternoon hours, primarily during daily commute. 

%% file: evaluation.tex
\section{Evaluation}
\label{section:eval}

Our study of outdoor walking surfaces aims to answer the following questions:

\begin{itemize}
    \item Can paving materials be classified via images captured on a mobile phone?
    \item Can we detect when a pedestrian transitions from sidewalk to street?
    \item How generalizable are models across cities?
    \item How does the camera-based technique compare with  existing techniques, such as shoe-sensing?
    \item How do various texture descriptors compare for diverse set of materials?
\end{itemize}

\begin{figure}[t]
    \centering
        \includegraphics[width=\linewidth]{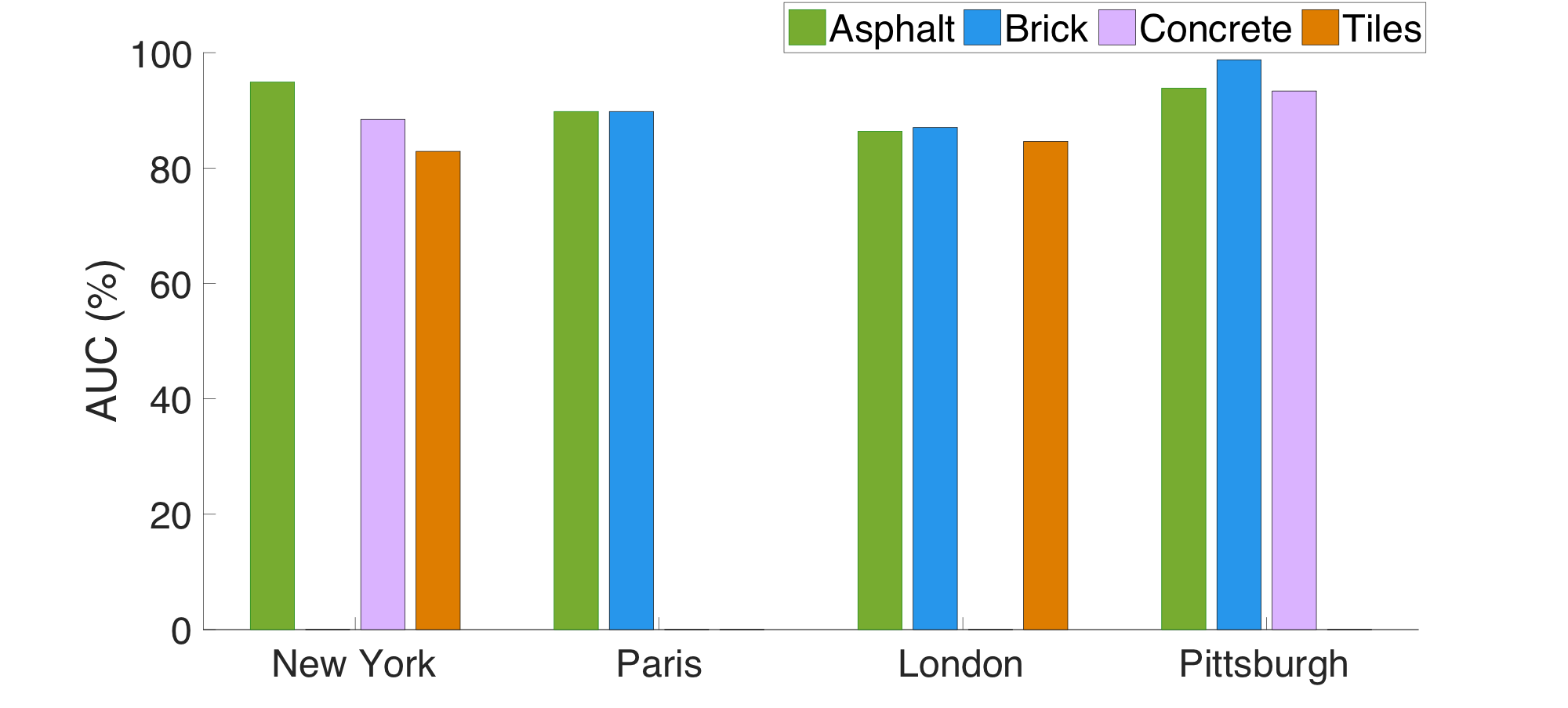}
        \caption{City-wise material classification.}
        \label{fig:material}
\end{figure} 

\subsection{Distinguishing paving materials on streets and sidewalks}

The future of navigation in autonomous cars and wheel-chairs, and large scale smart city data services relies heavily on the ability of cameras in detecting and distinguishing our environment, which includes paved surfaces. 
It is therefore pertinent to quantify how these materials are perceived by mobile cameras in different locations and varying weather and lighting conditions.
The choice of materials used for paving streets and sidewalks vary vastly, as can be seen in Table~\ref{tab:dataset}. There are many factors that determine this choice, primarily for maintenance and strength for expected traffic flow. 

In the United States for example, streets are commonly paved with asphalt, with more than 94\% of paved roads made of asphalt~\cite{asphalt}. Asphalt is hard and durable, and it is easy to replace damaged or broken asphalt. Concrete, often used for sidewalks, is installed in the form of stiff solid slabs that are prone to cracking and breaking. Brick, however expensive than concrete, is another commonly used material for sidewalk paving~\cite{concrete}. For the purpose of our classification, we combine clay brick and concrete brick under the category of bricks, also not considering the tessellation, or type of brick laying. 
In Paris, our dataset from two volunteers in different parts of the city, comprised primarily of asphalt streets and sidewalks. We also encountered several brick street crossings. We observed that in many of these places, concrete was still used to make curbs that separated street and sidewalk.
In London, sidewalks were commonly tiled or made of bricks. We use the multiclass classifier described in Section~\ref{sec:features}, to classify the materials in each city. For each material, we split the data into 3 parts, and use 2 parts for training and one for testing. During both, training and testing, we use the same number of observations from each class. The results are shown in Figure~\ref{fig:material}. Performance on the training data is not a good indicator of classifier performance on unseen data, therefore these results were obtained on an unseen test data, which was not used during the training phase. As a metric for classifier performance, we  use the area under the ROC curve (AUC). We believe it provides a more comprehensive understanding of classifier performance than overall accuracy because there is no hard threshold.  It can be seen that we can identify any material in our dataset with at least $85\%$ accuracy, with most  materials having a higher than $90\%$ AUC. We notice that our accuracy for tiles is lower than other materials. This is likely because we made simplifying assumptions during data labeling when the material of the tiles was not recognizable due to blurring caused by camera motion. Thus, we demonstrate that texture features can be  used for accurate material recognition in uncontrolled urban environments.

\begin{figure}[t]
    \centering
        \includegraphics[width=\linewidth]{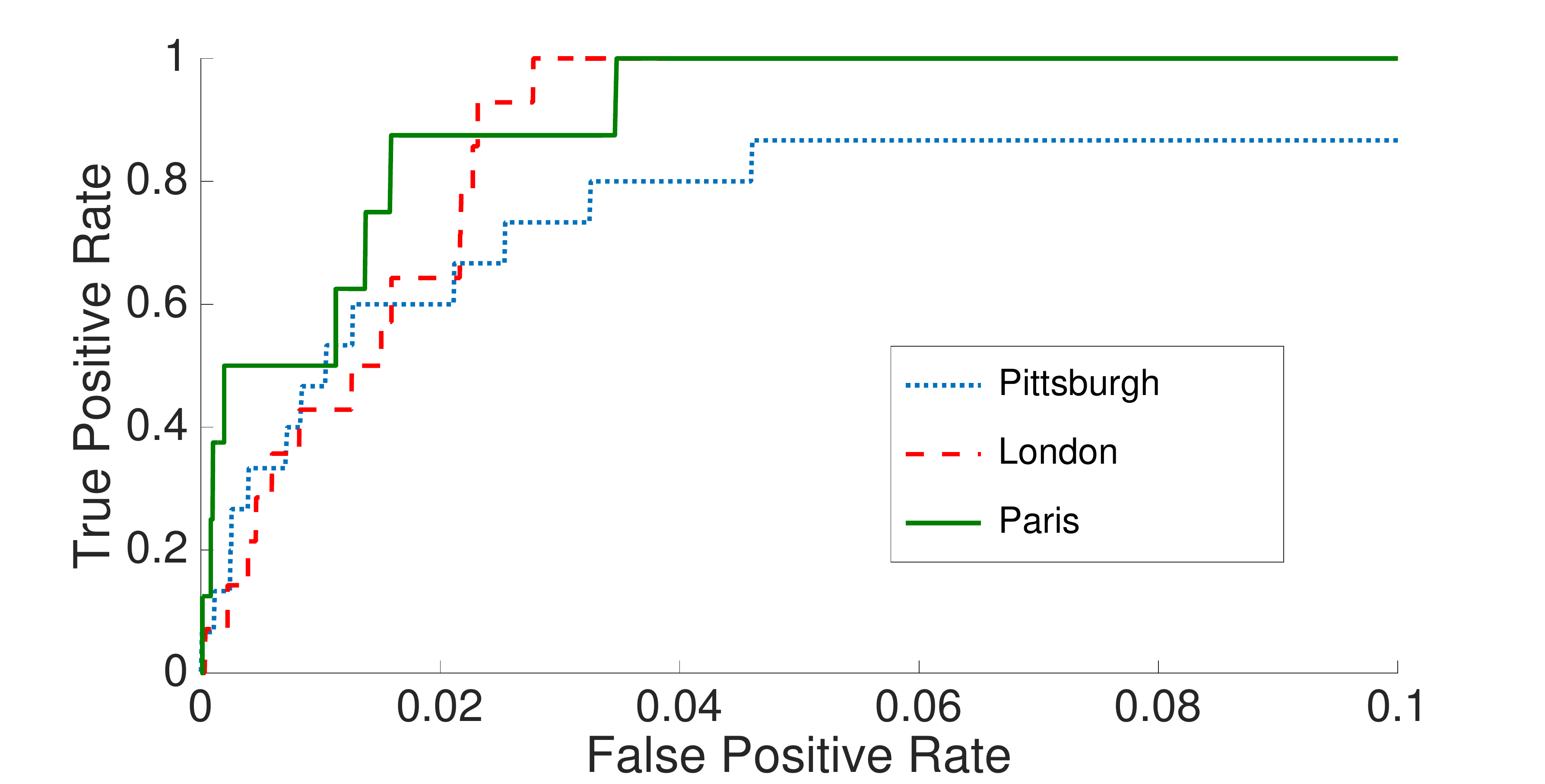}
        \caption{Entrance Detection Performance.}
        \label{fig:roc}
\end{figure}

\subsection{Street Entrance Detection}

We evaluate the proposed 
technique by detecting transitions from sidewalk to street, and their timeliness. 
We assign detection windows around the actual entrance. If a detection occurs in the specified window, it is considered correct. 
This window helps us evaluate the latency of detections, which in turn helps us estimate the usefulness of the warnings thus generated. We consider detections that occur within $2$ seconds before a pedestrian enters the street and at most one second after entering the street, to be useful in alerting the user to pay attention.
Figure~\ref{fig:roc} shows detection results that were evaluated on data collected in London, Paris, and Pittsburgh. We chose these three locations because of their similarity in data collection means, i.e. using a smartphone, as opposed to a GoPro used in the New York city dataset. Paris dataset comprises asphalt sidewalks and streets. Therefore entrance detection relies on detecting the transition curb or ramp accurately. We see that for our test walking trials, we can attain $100\%$ detection for London and $95\%$ for Paris. The lower performance in Paris is likely due to the fact that transitions between asphalt sidewalks and streets are not always marked by curbs. 
We see that for our test walking trials, we can attain $100\%$ detection for London  and Paris, and $85\%$ for Pittsburgh. The lower performance in Pittsburgh is likely due to the fact that street and sidewalk were commonly made of the same material, and transitions are harder to detect.
We obtain scores, for transition and not transition, for each observation, and determine the class by varying the score difference threshold from 0 to 1 in steps of 0.01. Based on the true and false positives determined, we plot the Receiver Operating Characteristic (ROC) that marks the true positive rate against the false positive rate, at each threshold. At a small sampling rate of 6 frames per second, we can detect greater than $90\%$ of entrance events, as the pedestrian enters the street, with less than $2.5\%$ false positives. This illustrates the timeliness of the detections.
We also find that the algorithm performance is unaffected by the camera tilt angle, if some part of the walking path features in the camera view.

\begin{figure}[t]
    \centering
        \includegraphics[width=0.4\textwidth]{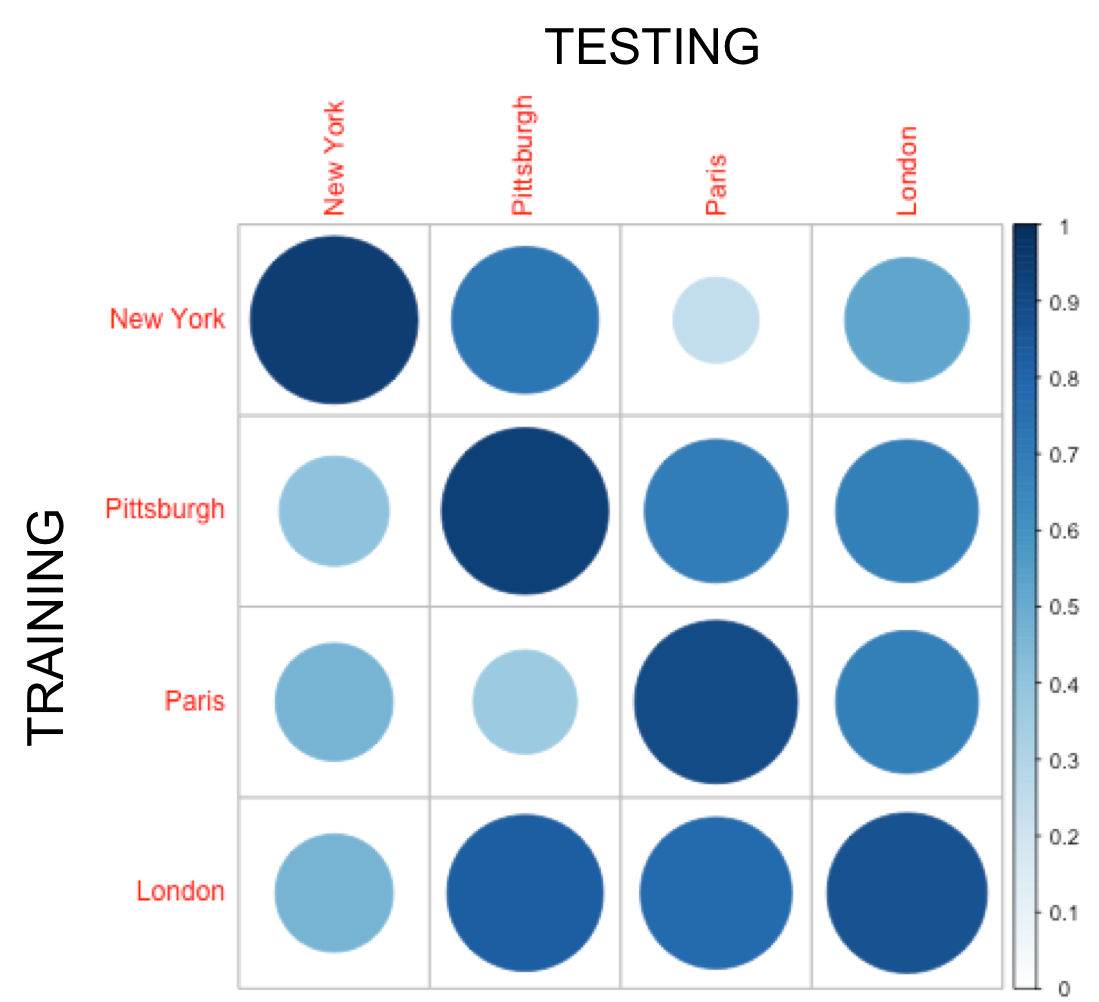}
        \caption{Cross-city asphalt classification.}
        \label{fig:crosscity}
\end{figure}

\subsection{Cross-city learning}

In the previous section, we obtained the results by training and testing on the data obtained in the same city. Given the overlap in the paving materials used across cities, we are interested in investigating the generalizability of models trained on each city. To realize this, we used models trained on  one city, to classify test data from another city. Since asphalt is the common material among all cities, we conduct these experiments for classifying asphalt. Figure~\ref{fig:crosscity} shows the results of our cross-city asphalt classification. It is clear that each city performs best with a model trained on the same city, which is expected in a machine-learning setting. Additionally, some cities have very strong similarity in the materials, for example a model trained in London performs very well in Pittsburgh and Paris ($>0.8$). Similarly, models trained in Pittsburgh and Paris perform very well in London. However, the low correlation with data in New York City is likely due to the fact that New York city data was collected using a GoPro, while the other data was collected using smartphones. The lower right $3 \times 3$ cells in Figure~\ref{fig:crosscity} indicates that almost all models collected using smartphones are easily usable in other cities. The key takeaway from this analysis is that when adding new cities in the mix, one need not train a model from scratch, and techniques such as transfer learning can be deployed to quickly converge the model to optimum performance.

\subsection{Comparing Feature Descriptors}

\begin{figure}[t]
    \centering
        \includegraphics[height=0.5\columnwidth]{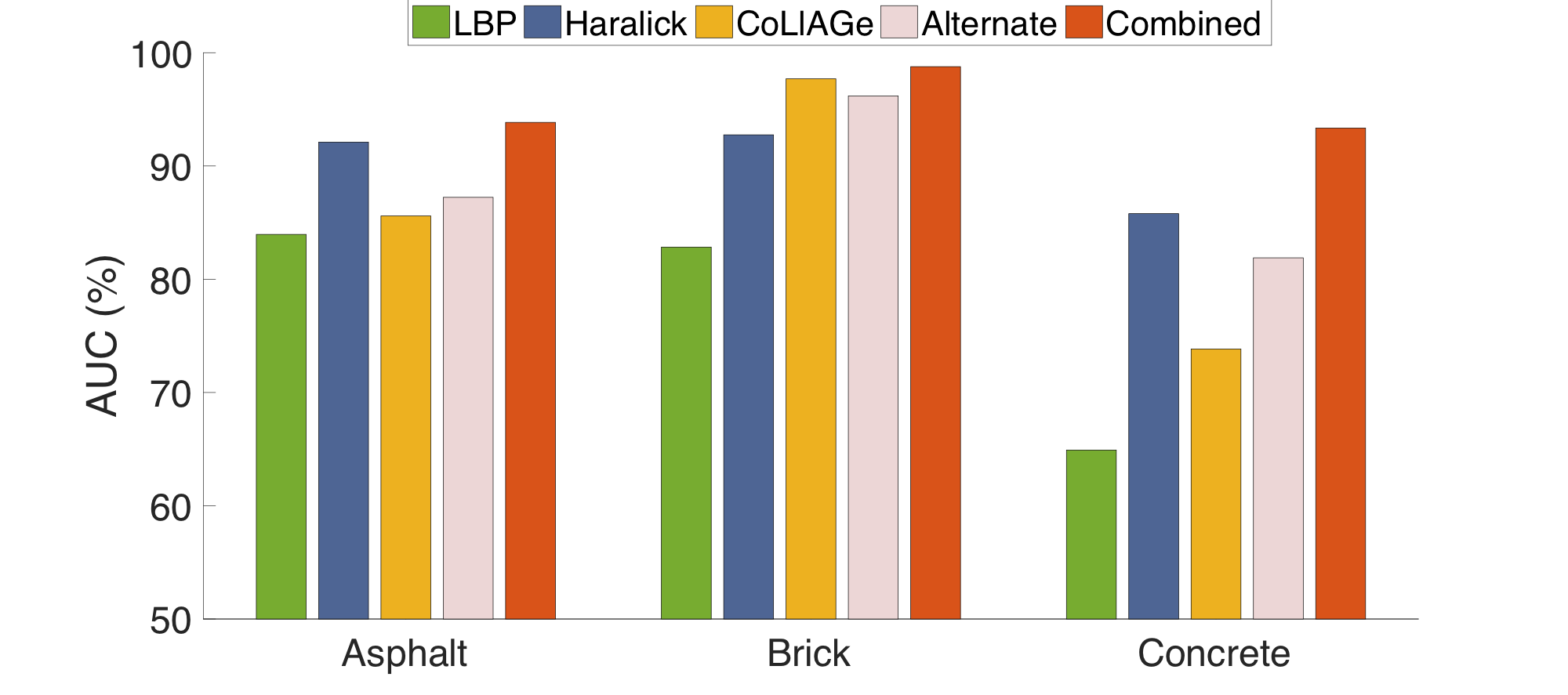}
        \caption{Comparison of feature descriptors based on classification performance. Data from Pittsburgh.}
        \label{fig:featurecomparison}
\end{figure}

The feature space $\mathcal{FS}$ presented in Section~\ref{sec:features} was defined after experimenting with several texture descriptors and quantifying their performance across various materials. Figure~\ref{fig:featurecomparison} shows the classifier performance obtained by using different feature descriptors. It is important to note that these features exhibit dissimilar performance for classifying different materials. For example, Local Binary Patterns (LBP) perform worse than all other features for all materials. While Haralick features perform well for all materials. Our combination of  alternate representations, haralick and CoLlAGe features is seen to give the best performance for all materials.

\subsection{Micro Benchmarks}

We implemented our algorithm on a Nexus 5X Android device using the OpenCV library~\cite{opencv} and JNI framework. The classifier was trained offline using OpenCV Support Vector Machine~\cite{svm} implementation, and the model was exported to the smartphone. 
This implementation was trained and tested on the data collected around our lab. With as few as $6$ frames per second we can obtain similar performance as presented in section~\ref{section:eval}. It takes approximately $62$ ms per frame for feature computation. Over an hour of continuous operation, the application consumes less than $5\%$ of battery charge. We consume very little power because instead of capturing videos at higher frame rates, we capture images only periodically. Moreover, the system is turned on only when the user is walking outdoors while also using the phone. We detect phone use by the status of the display. If the display is off, we assume that the user is not distracted by the phone, and is more aware of the surroundings. To conserve the power consumed by the display during camera, we reduce the display size to quarter of the screen when capturing the image.

\section {Comparison with gradient profiling }

In our previous work, we proposed shoe sensing based gradient profiling approach to detect sidewalk-street transitions. This approach detects roadway features such as ramps and curbs in crowded urban environments.
To quantitatively analyze the performance of our camera-based approach, we compare it to the shoe-based gradient profiling approach in LookUp!~\cite{LookUp}. 

LookUp was evaluated in two different urban environments. The first test site was Manhattan in New York City. The experiments were performed near Times Square, which is one of the world's busiest pedestrian intersections. 
The second test location was the the European city of Turin, in Italy. Of these, we use the camera data collected during LookUp experiments in Manhattan, New York.

We discuss the results from LookUp briefly.
First off, crossing detection algorithm is evaluated for delay and detection performance. This evaluation is carried out for the Manhattan  and Turin testbeds. These results establish that the crossing detection algorithm has very low false positives for a high detection rate, even at locations that have completely different street designs. 
LookUp uses steps as the evaluation metric, which provides a comprehension of  time and distance. 
To understand the timeliness of event detections, the delay distribution of the detections is analyzed.
Maximum number of detections occur at the step right before the entrance, followed by the first step into the street. The highest density of detected events lies in the steps before the entrance.

\begin{figure}[t]
    \centering
        \includegraphics[height=0.6\columnwidth]{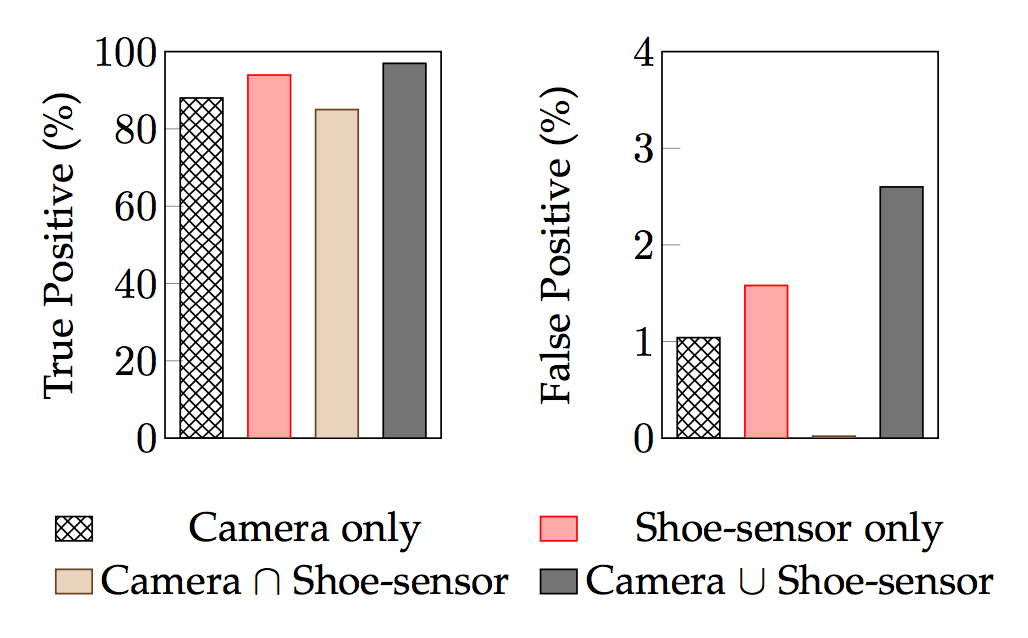}
       \caption{Comparison of Camera and Shoe-Sensing Approach. Data from New York.}
       \label{fig:comparison}
\end{figure}

\subsection{Comparison results}

Figure~\ref{fig:comparison} shows a comparison of the true positive rate and false positive rate from both the systems. It is important to remember that true positives are the correct detection of street entrance events, while false positives denote incorrect street detections. 
The camera system alone had a detection rate of $88\%$ compared to $94\%$ of the shoe-sensing. $85\%$ detections were common among both (intersection), while together (union), they exhibit a true positive rate of $97\%$. Of the total false positive rate of $2.6\%$, almost none were common among the two approaches. $1.5\%$ were caused by shoe sensors and $1.1\%$ by the camera approach. 
This reveals that while the gradient profiling approach has better performance, the camera-based approach may be slightly more resilient to false positives. Moreover, the absence of common false positives denotes that these two approaches are complimentary to each other, and thus can be potentially combined to formulate a robust system.


%% file: discussion.tex
\section{Discussion}

Accurately distinguishing materials in urban environments is a challenging problem due to the apparent similarities in their appearances. We have presented a mobile-camera based material classification approach, which unlike previous work, aims at recognizing texture rather than objects in camera's field of view. Through large-scale test data collected across cities, we have demonstrated that texture information can be used for distinguishing between materials in noisy outdoor environments. We developed a  street entrance detection algorithm based on the aforesaid material classification, as a system  for alerting distracted pedestrians when they enter the street.

The unique dataset from a pedestrian's perspective is one of our significant contributions. However, the data was labeled manually, and certain simplifying assumptions were made based on the pattern of the paving. For example, concrete bricks and clay bricks were both labeled in the broader bricks category. Similarly, asphalt includes both, streets with and without painted crosswalk. Sometimes, due to blurriness caused by motion, it is hard to exactly identify the material. The aforesaid simplifications help us narrow down the target groups for classification, and reduce complications. 

Camera sensing approaches, in general, are prone to lighting conditions, and can be severely impacted by ambient noise, such as bright lights. Additionally, real world environments are cluttered and the presence of unexpected objects can influence the algorithm performance. Like any vision-based technique, the performance of TerraFirma during night depends heavily on the presence of street lights. In the absence of street lights, this performance deteriorate. While the present data was collected mostly during the day, we will gather data during night time, and explore TerraFirma's performance.

Camera sensing is also vulnerable to obstruction by the user's hand during texting, which is more likely when the phone is being used in the landscape mode. 
Recently, with the advent of deep learning techniques, the performance of the system may be significantly improved, but our system is targeted at mobile cameras with meager resources. Considering the limitations in computation power and memory, and the rich image data, deep neural networks can be computationally intensive.

We contrast our approach to earlier approaches of smartphone-based sensors and shoe mounted inertial sensors. In terms of performance, we site our work in between these two approaches. In urban environments, camera sensing yields better results compared to other sensors on the smartphone, such as GPS and inertial sensors. However, even with efficient texture analysis techniques, considerable amount of work would be needed to match the performance of the shoe-sensor. Cameras potentially capture a much richer feature space than an inertial sensor, which would lead one to believe that higher accuracy should be possible. However, it is challenging to extract this information from imagery and even a design based on the state-of-art texture algorithms cannot yet match the accuracy of the dedicated shoe sensor. Nonetheless, it can detect street entrances irrespective of the ramps or curbs, that the inertial technique relies on. Consequently, it can also be effective in scenarios that inertial ground profiling is impervious to, such as where street and sidewalk are at the same level with no palpable difference in gradient. Overall, we have demonstrated that our technique can provide useful information when dedicated sensors are not available or complement inertial sensing approaches in mitigating false positives, and designing a robust system. 
This work is a demonstration of the feasibility of performing fine-grained pixel-based texture recognition on mobile cameras. One can further improve the performance by employing sophisticated vision techniques that can reduce blur and compensate for changing light conditions. 

%% file: conclusion.tex
\section{Conclusion}

To address the concerns surrounding heightened pedestrian risk, we explored the potential of smartphone-based camera sensing to identify paving materials in urban environments, and to detect sidewalk-street transitions. 
We collected in-the-wild walking data across complex metropolitan environments - New York, Paris, London, and Pittsburgh. This outdoor uncontrolled dataset will be released for public use, and is currently available upon request.
We show through experiments across four major cities of the world that texture analysis techniques can be effectively used for such classification. For material classification, our results show encouraging classification accuracy of more than $90\%$ for asphalt, brick, and concrete. We also evaluated our algorithm across cities, and achieve high accuracy for data collected on similar types of cameras, such as smartphones. 
Our entrance detection results show encouraging detection rates of $90\%$ with less than $3\%$ false positives. 
We demonstrate through an Android implementation that with lower frame capture rates, our system can be used favorably during routine outdoor walking sessions. 
Overall, by demonstrating that mobile cameras can be used for texture recognition and material classification in outdoor cluttered urban environments, we believe that we have introduced new avenues in mobile camera sensing.